\documentclass[11pt]{article}

\usepackage[final]{acl}

\usepackage{times}
\usepackage{latexsym}

\usepackage[T1]{fontenc}

\usepackage[utf8]{inputenc}

\usepackage{microtype}

\usepackage{inconsolata}

\usepackage{graphicx}
\usepackage{booktabs}
\usepackage{tikz}
\usepackage[table]{xcolor}
\usepackage{amsmath}
\usepackage{subcaption}
\usepackage{amsfonts}
\usepackage{pifont}
\usepackage{booktabs}
\usepackage{tabularx}
\usepackage{amssymb}
\usepackage{framed}
\usepackage{placeins}
\usepackage{makecell}
\usepackage[most]{tcolorbox}

\newtcolorbox{promptbox}[1][]{
  colback=white,
  colframe=black!55,
  boxrule=0.4pt,
  arc=1pt,
  left=4pt, right=4pt,
  top=4pt, bottom=4pt,
  #1
}

\newcommand{\cmark}{\textcolor{green!55!black}{\ding{51}}}
\newcommand{\xmark}{\textcolor{red!60!black}{\ding{55}}}
\newcolumntype{C}{>{\centering\arraybackslash}X}

\usetikzlibrary{decorations.pathreplacing,positioning,calc,arrows.meta,shapes,fit,backgrounds,bending}
\definecolor{kbblue}{RGB}{40,40,160}
\definecolor{wing}{RGB}{30,120,40}
\definecolor{setcol}{RGB}{30,70,170}
\definecolor{parcol}{RGB}{200,110,0}

\newcommand{\ci}[2]{$#1_{\pm#2}$}
\newcommand{\cibf}[2]{$\mathbf{#1}_{\pm#2}$}
\newif\ifshowstatus
\showstatustrue   

\newcommand{\done}[1]{#1}

\newcommand{\ment}[1]{\textbf{[#1]}}
\newcommand{\dec}[1]{\textbf{#1}}

\title{When Does Reasoning Help Entity Disambiguation?\\Small LLMs In-Domain and Under Shift}

\author{Samy Haffoudhi, Fabian M. Suchanek, and Nils Holzenberger \\
        Télécom Paris, Institut Polytechnique de Paris \\
        \texttt{\{samy.haffoudhi, fabian.suchanek, nils.holzenberger\}@telecom-paris.fr}}

\begin{document}
\maketitle
\begin{abstract}
Entity disambiguation (mapping ambiguous mentions in text to entities in a knowledge base) is a foundational step in tasks such as knowledge graph construction, question answering, and information extraction. Recent work has applied LLMs to this problem, achieving state-of-the-art performance. Notably, reasoning LLMs have been shown to perform entity disambiguation in a true zero-shot fashion, generalizing across knowledge bases and domains. However, it is unclear whether this capability survives, or remains competitive, with smaller LLMs, where entity disambiguation has typically relied on reranking rather than reasoning. 
We ask when reasoning is worth its cost at small scale. We compare
pointwise, listwise, and reasoning-based candidate-selection components within a classical coarse-to-fine entity disambiguation pipeline, across the classical, few-shot and true zero-shot settings.
The reasoning paradigm is trained with distillation and reinforcement learning recipes.
The benefit of reasoning grows with domain shift: negligible in-domain, decisive in the true zero-shot setting.
We show that two distinct factors are at work: the reranker's uncertainty over
the candidates locates \emph{where} the paradigms disagree, while shift
governs \emph{which} is right when they do.\footnote{\url{https://github.com/samyhaff/ed-study}}\textsuperscript{,}\footnote{\url{https://hf.co/collections/samyhaff/ed-study}}
\end{abstract}

\section{Introduction}

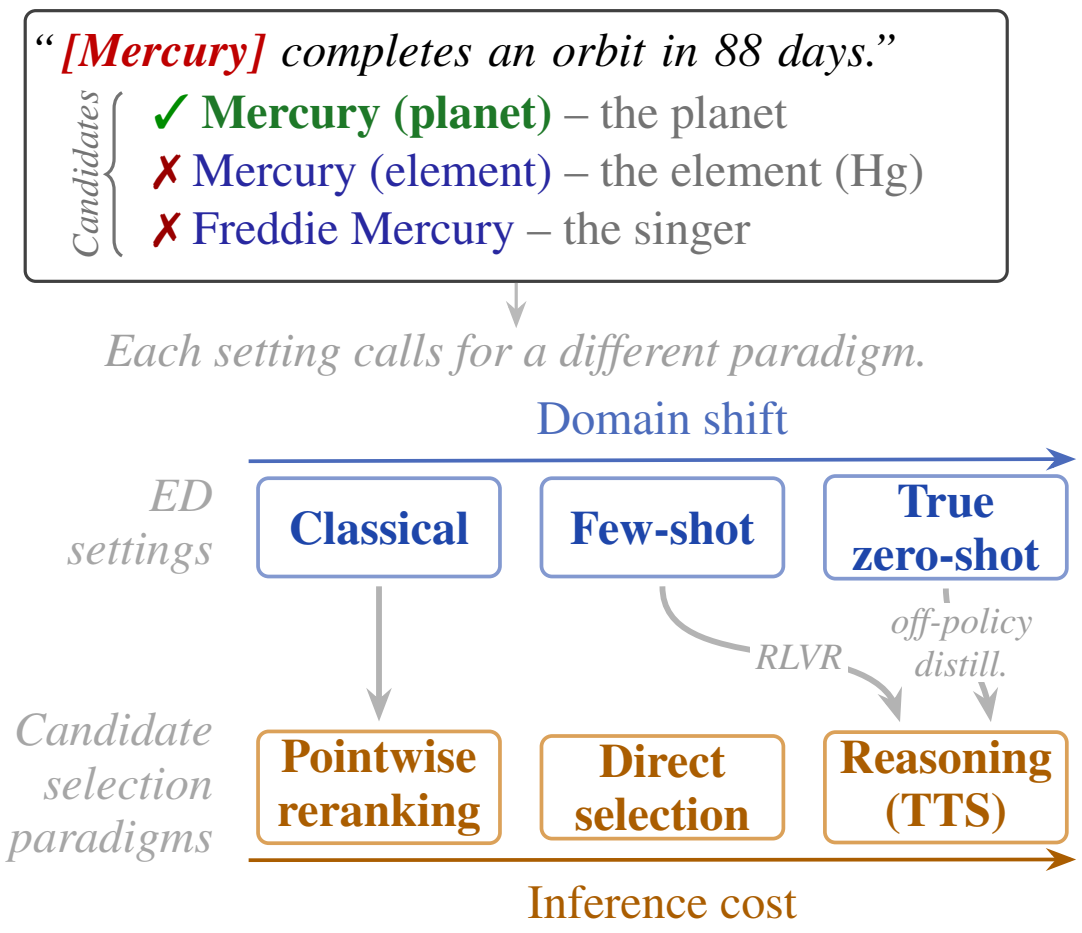
\begin{figure}[t]
\centering
\begin{tikzpicture}[
  font=\footnotesize,
  setbox/.style={
    draw=setcol!55, line width=0.7pt, rounded corners=2pt,
    align=center, text=setcol, font=\footnotesize\bfseries,
    minimum height=6.5mm, minimum width=16mm, inner sep=1.5pt},
  parbox/.style={
    draw=parcol!65, line width=0.7pt, rounded corners=2pt,
    align=center, text=parcol!85!black, font=\footnotesize\bfseries,
    minimum height=6.5mm, minimum width=16mm, inner sep=1.5pt},
  exframe/.style={
    draw=black!75, line width=0.6pt, rounded corners=2pt,
    fill=none, inner sep=4pt},
  extxt/.style={align=left, inner sep=0pt, text=black},
  candlbl/.style={rotate=90, font=\scriptsize\itshape,
    text=black!55, inner sep=0pt},
  cbrace/.style={decorate, decoration={brace, amplitude=3pt},
    draw=black!55, line width=0.6pt},
  axis/.style={-{Stealth[length=2mm]}, line width=0.7pt},
  feed/.style={-{Stealth[length=1.8mm]}, line width=0.6pt, gray!55},
  map/.style={-{Stealth[length=2.2mm]}, line width=1.1pt,
    draw=gray!60, shorten >=1.5pt, shorten <=1.5pt},
  maplbl/.style={font=\scriptsize\itshape, text=gray!75,
    fill=white, inner sep=1pt},
  grouplbl/.style={font=\small\itshape, text=gray!70}
]
\node[setbox] (c) {Classical};
\node[setbox, right=2.5mm of c] (f) {Few-shot};
\node[setbox, right=2.5mm of f] (z) {True\\zero-shot};
\node[parbox, below=10mm of c]  (rr)  {Pointwise\\reranking};
\node[parbox, right=2.5mm of rr] (ds) {Direct\\selection};
\node[parbox, right=2.5mm of ds] (tts){Reasoning\\(TTS)};
\node[grouplbl, left=1.5mm of c,  align=right] (lblS) {ED\\settings};
\node[grouplbl, left=1.5mm of rr, align=right] (lblP)
  {Candidate\\selection\\paradigms};
\node[extxt, anchor=south west] (excands)
  at ([xshift=7mm, yshift=13.5mm]lblS.north west) {%
  \cmark~{\color{wing}\textbf{Mercury (planet)}} \textcolor{black!55}{-- the planet}\\[1pt]
  \xmark~\textcolor{kbblue}{Mercury (element)} \textcolor{black!55}{-- the element (Hg)}\\[1pt]
  \xmark~\textcolor{kbblue}{Freddie Mercury} \textcolor{black!55}{-- the singer}%
};
\node[extxt, anchor=south west, text width=62mm] (exsent)
  at ([xshift=-7mm, yshift=3pt]excands.north west) {%
  \emph{``\,\textcolor{red!75!black}{\textbf{[Mercury]}} completes an orbit in 88 days.''}%
};
\draw[cbrace] ([xshift=-2mm]excands.south west)
           -- ([xshift=-2mm]excands.north west);
\node[candlbl, anchor=south] (lblCand)
  at ([xshift=-3.5mm]excands.west) {Candidates};
\node[exframe, fit=(exsent)(excands)(lblCand)] (ex) {};
\node[font=\footnotesize\itshape, text=gray!75, below=2mm of ex] (bridge)
  {Each setting calls for a different paradigm.};
\draw[feed] (ex.south) -- ++(0,-3mm);
\draw[map] (c.south) -- (rr.north);
\draw[map] (f.south) to[out=-90, in=90]
  node[maplbl, pos=0.55] {RLVR} ([xshift=-3mm]tts.north);
\draw[map] (z.south) to[out=-90, in=90]
  node[maplbl, pos=0.4, align=center] {off-policy\\distill.}
  ([xshift=3mm]tts.north);
\draw[axis, setcol!80] ($(c.north west)+(-0.5mm,1mm)$) -- ($(z.north east)+(0.5mm,1mm)$)
  node[midway, above=0.3mm, font=\small, text=setcol!80] {Domain shift};
\draw[axis, parcol!85!black] ($(rr.south west)+(-0.5mm,-1mm)$) -- ($(tts.south east)+(0.5mm,-1mm)$)
  node[midway, below=0.3mm, font=\small, text=parcol!85!black] {Inference cost};
\end{tikzpicture}
\caption{Given an ambiguous mention (\textbf{[Mercury]}) in context, entity
disambiguation maps it to the correct KB entity (the planet). We study which
candidate-selection paradigm (bottom, ordered by inference cost) best suits
each ED setting (top, ordered by domain shift), for small LLMs (0.6--4B).
Arrows give the mapping we find; labels give the training recipe for the
reasoning paradigm.}
\label{fig:overview}
\vspace{-\baselineskip}
\end{figure}

Entity Disambiguation (ED) is the task of mapping ambiguous mentions of entities in natural language text
to reference entities in a knowledge base (KB), each described by a name and a short description.
For instance, in ``\textbf{[Mercury]} completes an orbit in 88 days'', the mention Mercury must be resolved to the planet
rather than to the chemical element or to the singer (Figure~\ref{fig:overview}).

ED is part of the Entity Linking task, where mentions also have to be identified in the text~\cite{sevgiliNeuralEntityLinking2022,decaoAutoregressiveEntityRetrieval2021,haffoudhiLELAEndtoendLLMbased2026a} and is an important preprocessing step for Information Extraction~\cite{machine-knowledge-survey}, knowledge-based question answering~\cite{liEfficientOnePassEndtoEnd2020a},
knowledge graph completion~\cite{jiKnowledgeBasePopulation2011} and grounding language models in KBs~\cite{crespinKARLAKnowledgebaseAugmented2026}.
ED can be performed in a classical, few-shot, zero-shot, or true zero-shot setting, based on how close the training data is to the target~\cite{haffoudhiLELALLMBasedEntity2026a}. 

Coarse-to-fine retrieve-and-rerank is the most widely adopted paradigm for ED~\cite{wuScalableZeroshotEntity2020}: candidates are first retrieved, then scored against the context by a reranker.
Traditionally, these pipelines relied on BERT-based Pretrained Language Models~\cite{wuScalableZeroshotEntity2020}, and
recently, LLMs have pushed the ED state-of-the-art~\cite{haffoudhiLELALLMBasedEntity2026a,zhouGenDeciderIntegratingNone2024,boscariolEvaluationLLMsLongtail2025,liLeveragingPowerLarge2025,yeLLMEntityDisambiguator2025}.
In particular, Reasoning LLMs enable true zero-shot entity linking~\cite{haffoudhiLELALLMBasedEntity2026a}, generalizing across KBs and domains, by adding a third reasoning-based LLM selection step, receiving as input the pointwise-reranker's top-$k$ candidates.
However, these results were obtained with large models, and the long chain-of-thought-based approaches induce substantial inference cost over the classical two-stage pipeline. Hence, it is unclear whether reasoning-based selection remains viable, or even beneficial, at small scales, and how those different paradigms compare in the other ED settings. 
Meanwhile, a growing body of work studies how to elicit reasoning in language models through knowledge distillation and reinforcement learning~\cite{guoDeepSeekR1IncentivizesReasoning2025,muennighoffS1SimpleTesttime2025a,shaoDeepSeekMathPushingLimits2024,liLanguageModelsCan2025}, but whether such methods transfer to the ED task remains largely unexplored.
Two questions follow: (1) given a small LLM backbone, which selection paradigm — pointwise reranking, direct listwise selection, or reasoning-based selection — is best suited to each ED setting? (2) Can ED reasoning capabilities be transferred to small models, and which training paradigm — knowledge distillation or reinforcement learning — does so most effectively?
To answer these questions, we conduct a systematic study of small LLM backbones (0.6–4B) for ED across the classical, few-shot and true zero-shot\footnote{We omit plain zero-shot, as unseen test entities occur throughout, leaving domain shift as the only distinct challenge.} settings (see Figure~\ref{fig:overview}).
First, we determine which candidate selection paradigm maximizes performance in each ED setting, and at what inference cost. Second, we find that the more uncertain the reranker is over its top-1 prediction, the more it is worth resorting to a reasoning model. Finally, we show that domain shift matters because it drives that uncertainty up, and independently changes which candidate selection paradigm is right.

\section{Related Work}

\subsection{Entity Disambiguation}
ED is a core component of Entity Linking together with mention
detection~\cite{sevgiliNeuralEntityLinking2022,decaoAutoregressiveEntityRetrieval2021,haffoudhiLELAEndtoendLLMbased2026a}.
Classical supervised approaches are built on BERT-style encoders~\cite{gillickLearningDenseRepresentations2019,wuScalableZeroshotEntity2020},
while generative methods decode entity identifiers
autoregressively~\cite{decaoAutoregressiveEntityRetrieval2021}. A parallel line
studies ED under progressively weaker supervision: the \emph{zero-shot} setting,
where test entities are unseen but drawn from the training
domain~\cite{logeswaranZeroShotEntityLinking2019,tangBidirectionalMultiparagraphReading2021,xuReadandSelectFrameworkZeroshot2023,sunTransformationalBiencoderInDomain2022,zhouGenDeciderIntegratingNone2024,yaoZeroshotEntityLinking2020}, the \emph{few-shot} setting,
where only a handful of target-domain mentions are
available~\cite{liEffectiveFewShotNamed2022,liuOneNetFineTuningFree2024a,chenRevisitingSparseRetrieval2023}, and the \emph{true zero-shot}
setting, where neither entities nor domain are seen during
training~\cite{haffoudhiLELALLMBasedEntity2026a}. We adopt this taxonomy
(Table~\ref{tab:el_settings_details_here}) as the axis along which we compare candidate
selection paradigms.

\subsection{LLM-based Entity Disambiguation}

While retrieve-and-rerank dominate ED, 
LLMs have recently been applied to the candidate selection stage~\cite{zhouGenDeciderIntegratingNone2024,yeLLMEntityDisambiguator2025,shlykBeLinkBiomedicalEntity2026a}, 
either scoring candidates pointwise, selecting from a shortlist in a listwise
multiple-choice formulation, or, most recently, reasoning explicitly before
selecting. In particular, \citet{haffoudhiLELALLMBasedEntity2026a} show that a
reasoning-based selection step enables
\emph{true zero-shot} ED. These gains, however, were obtained with large models and incur 
substantial inference cost. 
\citet{liLeveragingPowerLarge2025} and concurrent work
SelectED~\citep{ruckerSelectEDSelectiveLLM2026} selectively invoke LLMs
based on mention difficulty or confidence; our analysis characterizes
\emph{when} reasoning justifies its added cost.

\subsection{Eliciting Reasoning in Small LLMs}

Two families of methods elicit reasoning in language models. \emph{Distillation}~\cite{hintonDistillingKnowledgeNeural2015,kimSequenceLevelKnowledgeDistillation2016a}
transfers a teacher's behavior into a student, either off-policy~\cite{guoDeepSeekR1IncentivizesReasoning2025,guhaOpenThoughtsDataRecipes2025a,liLanguageModelsCan2025}
or on-policy~\cite{yangQwen3TechnicalReport2025,agarwalOnPolicyDistillationLanguage2023,guMiniLLMKnowledgeDistillation2023}.
\emph{Self-Distillation} trains on-policy with the teacher being the student model with privileged information~\cite{shenfeldSelfDistillationEnablesContinual2026a,zhaoSelfDistilledReasonerOnPolicy2026a}.
\emph{Reinforcement learning from verifiable rewards} (RLVR), typically with
GRPO~\cite{shaoDeepSeekMathPushingLimits2024}, 
instead optimizes a reward function directly. 
Analyses of these paradigms suggest (1)~that RLVR sharpens the sampling of solutions that the base model can already reach, while SFT introduces new capability~\cite{yueDoesReinforcementLearning2025a,kimReinforcementLearningVs2025} and (2)~that 
SFT memorizes while RLVR generalizes~\cite{chuSFTMemorizesRL2025a,zhangGeneralizationSpectrumChromatographic2026,huanDoesMathReasoning2026}.
While these methods are well studied on
mathematical and coding reasoning~\cite{liAchievingGoldMedalLevelOlympiad2026,xuVibeThinker3BExploringFrontier2026}, 
whether they transfer to ED, and how they compare across model scale and domain shift, remains unexplored.

\section{Methodology}

\subsection{Task Definition and ED Settings}

Following~\citet{haffoudhiLELALLMBasedEntity2026a}, ED is defined using 4 random variables: (1)~the domain $D$ (e.g. News, Biomedical etc), (2)~the set of relevant entities $E$, (3)~the text document $T$, and (4)~the set of mentions $M$. It is posited that
the set of relevant entities $E$ only depends on the domain $D$;
the document $T$ only depends on the domain $D$;
and the set of mentions $M$ only depends on the document $T$ and the set of entities $E$.

\noindent The task of ED is that of linking the mentions $m \in M$ in a text document $T$ to their entities in $E$. We distinguish the following settings (Table~\ref{tab:el_settings_details_here}):
In \emph{classical ED}, one has access to documents sampled from a distribution conditioned on the test entities and test domain. The \emph{few-shot setting} shares the same structure as classical ED, but assumes that only $k \ll |E|$ mentions are observed during training.
In \emph{zero-shot ED}, the entities are distinct between train and test, but are sampled from the same domain. In the \emph{true zero-shot setting}, none of the distributions are conditioned on the test domain.

\begin{table}
\centering
\small
\setlength{\tabcolsep}{5pt}
\begin{tabular}{@{}l ccc@{}}
\toprule
Setting & Domain $D$ & Entities $E$ & Docs $T$ \\
\midrule
Classical      & $=d_{\text{test}}$ & $=e_{\text{test}}$ & $\sim T\!\mid\! d_{\text{test}}$ \\
Few-shot       & $=d_{\text{test}}$ & $=e_{\text{test}}$ & $\sim T\!\mid\! d_{\text{test}}$ \\
Zero-shot      & $=d_{\text{test}}$ & $\sim E\!\mid\! d_{\text{test}}$ & $\sim T\!\mid\! d_{\text{test}}$ \\
True zero-shot & $\sim D$ & $\sim E\!\mid\! d_{\text{train}}$ & $\sim T\!\mid\! d_{\text{train}}$ \\
\bottomrule
\end{tabular}
\caption{ED settings by the training samples available from the test domain.
$d_{\text{train}},d_{\text{test}}$ denote train/test domains (likewise $e$ for
entities); a cell shows how the train-time variable relates to the test
distribution ($=$: identical; $\sim$: sampled). Mentions, always conditioned on
the available document and entities, are omitted.}
\label{tab:el_settings_details_here}
\end{table}

\subsection{LLM-based Entity Disambiguation Approaches}
\label{sec:approaches}

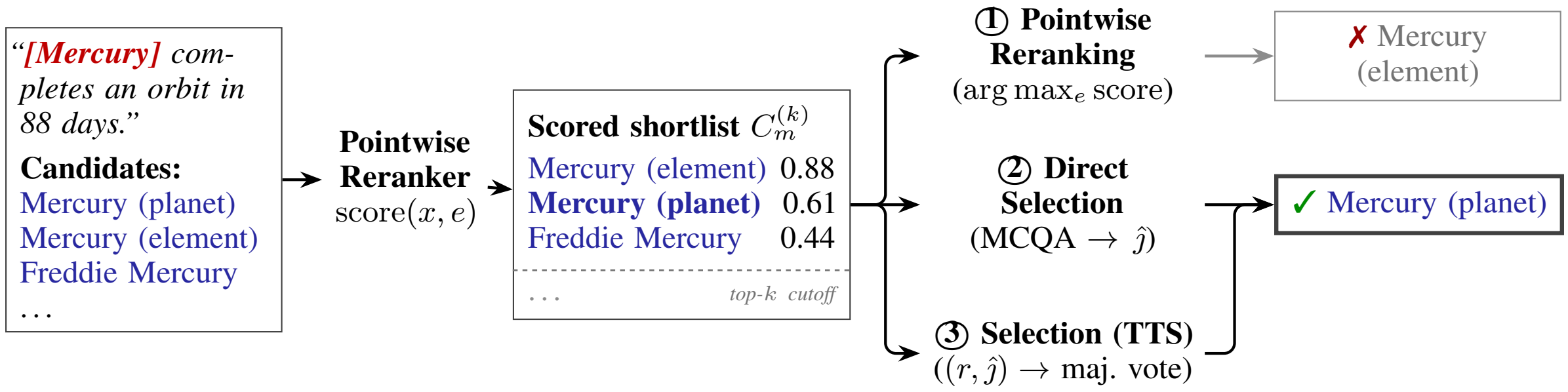
\begin{figure*}
\centering
\begin{tikzpicture}[
  >={Stealth[length=2.2mm]},
  font=\footnotesize,
  cell/.style={rectangle, draw, align=left, inner sep=4pt},
  input/.style={cell, draw=black!75, text width=2.5cm},
  data/.style={cell, draw=black!75, text width=3.1cm},
  proc/.style={align=center, font=\footnotesize\bfseries, text width=2.6cm},
  fin/.style={cell, draw=black!75, very thick, align=center, text width=2.6cm},
  finbad/.style={cell, draw=black!45, align=center, text width=2.6cm},
  arr/.style={->,thick, rounded corners=3pt},
  badarr/.style={->,thick, rounded corners=3pt, draw=black!45},
  extxt/.style={align=left, inner sep=0pt, text=black},
  slframe/.style={draw=black!75, rectangle, fill=none, inner sep=4pt},
  cutline/.style={dashed, dash pattern=on 1.4pt off 1.4pt,
                  draw=black!55, line width=0.5pt},
  badge/.style={draw, circle, fill=white, font=\tiny, inner sep=0.5pt},
]
\node[input] (in) {%
  \emph{``\textcolor{red!75!black}{\textbf{[Mercury]}} completes an orbit in 88 days.''}\\[3pt]
  \textbf{Candidates:}\\
  \textcolor{kbblue}{Mercury (planet)}\\
  \textcolor{kbblue}{Mercury (element)}\\
  \textcolor{kbblue}{Freddie Mercury}\ \dots%
};
\node[proc, right=0.4cm of in, text width=1.4cm] (rr)
  {Pointwise\\ Reranker\\ \footnotesize\mdseries $\operatorname{score}(x,e)$};
\draw[arr] (in) -- (rr);
\node[extxt, right=0.4cm of rr, text width=3.1cm] (slkeep) {%
  \textbf{Scored shortlist} $C_m^{(k)}$\\[2pt]
  {\color{kbblue}Mercury (element)}\hfill 0.88\\
  {\color{kbblue}\textbf{Mercury (planet)}}\hfill 0.61\\
  {\color{kbblue}Freddie Mercury}\hfill 0.44%
};
\node[extxt, text width=3.1cm, anchor=north west] (slcut)
  at ([yshift=-3mm]slkeep.south west) {%
  {\color{black!45}\dots}\hfill{\itshape\color{black!55}\tiny top-$k$ cutoff}%
};
\coordinate (cuty) at ($(slkeep.south)!0.5!(slcut.north)$);
\node[slframe, fit=(slkeep)(slcut)] (sl) {};
\draw[cutline] (sl.west|-cuty) -- (sl.east|-cuty);
\draw[arr] (rr) -- (sl);
\node[proc, right=0.7cm of sl, yshift=1.5cm] (b1)
  {\textcircled{1} Pointwise Reranking\\ \footnotesize\mdseries($\arg\max_e \operatorname{score}$)};
\node[proc, right=0.7cm of sl] (b2)
  {\textcircled{2} Direct Selection\\ \footnotesize\mdseries(MCQA $\to\hat\jmath$)};
\node[proc, right=0.7cm of sl, yshift=-1.5cm] (b3)
  {\textcircled{3} Selection (TTS)\\ \footnotesize\mdseries($(r,\hat\jmath)\to$ maj.\ vote)};
\draw[arr] (sl.east) -- ++(0.35,0) |- (b1.west);
\draw[arr] (sl.east) -- ++(0.35,0) -- (b2.west);
\draw[arr] (sl.east) -- ++(0.35,0) |- (b3.west);
\node[finbad, right=0.7cm of b1] (finbad)
  {\xmark~\textcolor{black!55}{Mercury (element)}};
\node[fin, right=0.7cm of b2] (fin)
  {\cmark~\textcolor{kbblue}{Mercury (planet)}};
\draw[badarr] (b1.east) -- (finbad.west);
\draw[arr] (b2.east) -- (fin.west);
\draw[arr] (b3.east) -- ++(0.3,0) |- (fin.west);
\end{tikzpicture}
\caption{The three candidate-selection paradigms, left to right. A pointwise
reranker scores the retrieved candidates ($n_c$ passes) and keeps the top-$k$,
yielding the scored shortlist $C_m^{(k)}$. Here the reranker ranks the gold
entity \emph{second}. The three paradigms then differ:
\textcircled{1}~\textbf{Pointwise Reranking} takes the top-scored candidate via
$\arg\max$ and inherits the error; \textcircled{2}~\textbf{Direct Selection} is
fed the whole shortlist and returns an index $\hat\jmath$, so it can recover the
gold entity; \textcircled{3}~\textbf{Selection (TTS)} generates $n$
chain-of-thought completions $(r,\hat\jmath)$ and majority-votes.}
\label{fig:inference_methods}
\end{figure*}

Given the potentially millions of entities in a KB, ED is
typically tackled in two stages~\cite{wuScalableZeroshotEntity2020}:
\emph{candidate generation}, producing a set
$C(m,E)\subseteq E$ of $n_c$ candidates for a mention $m$; and \emph{candidate selection}, choosing an entity
$\hat{e}\in C(m,E)$. We fix the retrieval stage and compare three LLM-based selection paradigms, 
all sharing an initial pointwise reranking pass that produces a top-$k$
shortlist $C^{(k)}_m$~\cite{haffoudhiLELALLMBasedEntity2026a} (Figure~\ref{fig:inference_methods}).
Prompts are given in Appendix~\ref{sec:prompts}. Summaries of the
paradigm and training objective characteristics are provided in
Table~\ref{tab:paradigm-comparison} in the Appendix and Table~\ref{tab:training-comparison} 
(Appendix~\ref{app:objectives}).

\paragraph{Pointwise reranking.}
Each candidate is scored independently by an LLM-based reranker~\cite{zhangQwen3EmbeddingAdvancing2025}. We set the query to the context $T$ (with $m$ highlighted)
and the document $d$ to the title and description of a candidate $e$, and
compute
\begin{equation}
\mathrm{score}(T, e) = \sigma\!\big(\ell_{\text{yes}} - \ell_{\text{no}}\big),
\end{equation}
where $\ell_{\text{yes}},\ell_{\text{no}}$ are the logits for a
yes/no relevance judgment and $\sigma$ the logistic function. The
prediction is the arg-max over the full candidate set,
$\hat{e}=\arg\max_{e\in C(m,E)}\mathrm{score}(T,e)$. This stage requires
$n_c$ forward passes per mention and sees every candidate, but scores
them in isolation, without cross-candidate comparison.

\paragraph{Direct selection.}
The top-$k$ shortlist $C^{(k)}_m$ is passed to an instruction-tuned LLM
in a multiple-choice formulation: the prompt contains the context $T$
with $m$ highlighted and the $k$ candidates as an indexed list of
(name, description) pairs. The model emits a single index
$\hat{\jmath}\in\{1,\dots,k\}$ and we predict
$\hat{e}=C^{(k)}_m[\hat{\jmath}]$. Unlike reranking, the model compares
candidates jointly in one pass.

\paragraph{Test-time Scaling-based selection (TTS).}
This paradigm uses the same multiple-choice setup but generates an explicit chain-of-thought $r$
before its answer, producing a completion $y=(r,\hat{\jmath})$. At test
time we draw $n$ independent completions and aggregate their predicted
indices by majority vote\footnote{Ties are broken by sampled-completion order.}:
\begin{equation}
\hat{e}=C^{(k)}_m\!\Big[\arg\max_{j}\textstyle\sum_{l=1}^{n}
\mathbf{1}[\hat{\jmath}^{(l)}=j]\Big].
\end{equation}
Reasoning and self-consistency add cross-candidate deliberation
and test-time compute, at a cost of $n$ long-form generations per
mention.

\subsection{Training the Selectors}
\label{sec:training}

\paragraph{Reranking and direct selection.}
The pointwise reranker is trained with a binary cross-entropy loss over
its relevance score $\mathrm{score}(T,e)$, one example per candidate,
labelled positive iff $e$ is the gold entity. The direct selector is
trained by supervised fine-tuning (SFT): with $j^\ast$ the gold index in
the shortlist, we minimize the cross-entropy
$-\log p_\theta(j^\ast\mid T,m,C^{(k)}_m)$ on the answer token.

\paragraph{TTS-based selection.}
We compare:

\emph{RLVR.} We optimize the selector using GRPO~\cite{shaoDeepSeekMathPushingLimits2024},
with a binary correctness reward $R(y)=\mathbf{1}[\hat{\jmath}=j^\ast]$.
For each prompt we sample a group of $G$ completions and form
group-normalized advantages $A^{(g)}=(R(y^{(g)})-\mu)/\sigma$, where
$\mu,\sigma$ are the within-group reward mean and standard deviation.
The full objective is given in Appendix~\ref{app:grpo}.

\emph{Distillation.} Both distillation variants are instances of Generalized Knowledge
Distillation (GKD)~\cite{agarwalOnPolicyDistillationLanguage2023}, which
minimizes a divergence between teacher and student token distributions
over a mixture of teacher- and student-generated trajectories:
\begin{equation}
\begin{split}
\mathcal{L}_{\text{GKD}}(\theta,\lambda,\beta) =\;
& (1{-}\lambda)\,\mathbb{E}_{\pi_T}\!\big[D^{\beta}_{\text{JSD}}(p_T\|p_\theta)\big]\\
+\;& \lambda\,\mathbb{E}_{\pi_\theta}\!\big[D^{\beta}_{\text{JSD}}(p_T\|p_\theta)\big],
\end{split}
\label{eq:gkd}
\end{equation}
where $\lambda\in[0,1]$ controls the on-policiness and $D^{\beta}_{\text{JSD}}$
is the generalized Jensen--Shannon divergence, recovering the forward KL
$D_{\mathrm{KL}}(p_T\|p_\theta)$ as $\beta \to 0$ and the reverse KL
$D_{\mathrm{KL}}(p_\theta\|p_T)$ as $\beta \to 1$~\cite{huszarHowNotTrain2015} (Appendix~\ref{app:divergences}).
Our main comparison focuses on two natural endpoints. \emph{Off-policy distillation} $(\lambda,\beta){=}(0,0)$
trains the student on the teacher's traces under the forward KL.
\emph{On-policy distillation (OPD)} $(\lambda,\beta){=}(1,1)$
samples trajectories from the student and scores them with the teacher under the reverse KL~\cite{guMiniLLMKnowledgeDistillation2023}. 
The explicit forms of both terms are given in Appendix~\ref{app:gkd}.
We also ablate trajectory source and KL direction (Section~\ref{sec:analyses}).

\subsection{Training Data}
\label{sec:data}

All selectors are trained from a shared pool of mentions, each with its
gold entity $e^\ast$ and candidate set; a base reranker scores the
candidates and the top $k$ are kept as the shortlist $C^{(k)}$, shuffled
to remove position bias. From this pool we derive the training set. 
For \emph{reranking}, we flatten each shortlist into $k$
binary examples (positive iff the candidate is $e^\ast$);
for \emph{direct selection}, each example is a (context, shortlist,
gold-index) tuple. For the TTS paradigms, we roll out
$n_{\text{out}}$ completions per mention and filter them by correctness.
Both distillation variants share the same mention filter. Off-policy distillation then trains on those teacher traces directly; on-policy distillation instead rolls out the student on the same mentions and scores it with the teacher. \emph{RLVR} keeps
mentions whose student success rate lies within a band $[p_{\min}, p_{\max}]$, discarding the trivially easy and the unsolvable.

\section{Experiments}

\begin{figure}[t]
    \centering
    \includegraphics[width=.9\linewidth]{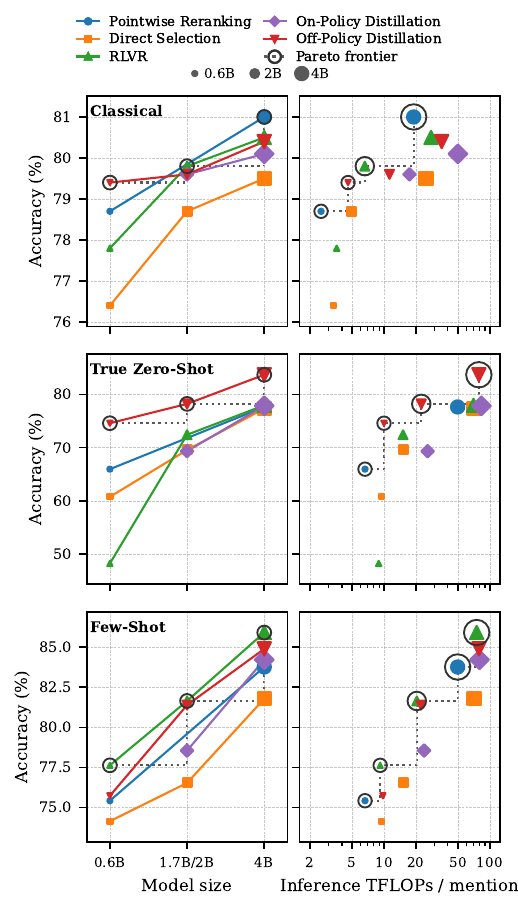}
    \caption{Accuracy vs.\ base model size (\textbf{left}) and vs.\ inference FLOPs per linked mention (\textbf{right}, log scale), per setting. FLOPs exclude retrieval.}
    \label{fig:results_summary}
\end{figure}

We report results across the three ED settings, along with a summary (Figure~\ref{fig:results_summary}). 

\subsection{Experimental Settings}

We experiment with the \texttt{Qwen3-0.6B}, \texttt{Qwen3-1.7B}, \texttt{Qwen3-4B} models~\cite{yangQwen3TechnicalReport2025}, the \texttt{Qwen3-Reranker-0.6B}
and \texttt{Qwen3-Reranker-4B} reranker variants~\cite{zhangQwen3EmbeddingAdvancing2025}\footnote{No 1.7B variant available.} and the \texttt{gemma-4-E2B-it} and \texttt{gemma-4-E4B-it}
models~\cite{teamGemma4Technical2026}, spanning two model families. 
In the results that follow, we group all of those models into three 0.6B, 1.7-2B and 4B groups for fair comparison.
We use the \texttt{gemma-4-26B-A4B-it} model as a teacher for
distillation\footnote{Here, for Qwen students,
off-policy distillation reduces to SFT on the gemma teacher's completions.} (see Appendix~\ref{sec:additional_results} for additional results).
We used the \texttt{trl}\footnote{\url{https://github.com/huggingface/trl}} library for training and 
\texttt{vllm}\footnote{\url{https://github.com/vllm-project/vllm}} for inference. 
We trained all models using LoRA~\cite{huLoRALowRankAdaptation2021a} and report training parameters in Appendix~\ref{app:training} and compute in Appendix~\ref{app:compute}.
Following \citet{haffoudhiLELALLMBasedEntity2026a}, we consider $k = 10$ for the reranking cutoff (see Appendix~\ref{sec:acck}) and sample $n = 10$ completions per mention for
majority voting and Pass@k metric computation (Appendix~\ref{sec:passk}). We set the maximum generation length to $2048$ tokens and 
use the recommended generation parameters for each model.
Each selection model is fed the shortlist of the largest reranker no
bigger than itself, so that no paradigm benefits from a larger upstream model.
We also report results from prior methods. In particular, we include LELA~\cite{haffoudhiLELALLMBasedEntity2026a}, 
the training-free LLM-based selector that our test-time-scaling paradigm builds on.

\subsection{The Classical Setting}
\label{sec:classical}

\paragraph{Benchmark} ZELDA \cite{milichZELDAComprehensiveBenchmark2023} groups established Wikipedia-based ED benchmarks.
It aims at allowing for a fair comparison of supervised methods by mapping all mentions to the same Wikidata KB and by providing a common
Wikipedia-based train set and candidate dictionary to be used across benchmarks and methods. 

\paragraph{Configuration.} We use the provided candidates and entity descriptions, and accuracy as the evaluation metric. 
We build our training pool from the ZELDA Wikipedia training split, which provides prior-based candidate dictionaries, and derive the per-paradigm training sets as described in Section~\ref{sec:data}, using ${k=10}$ for the reranked shortlist and ${n_{\text{out}}=10}$ rollouts per mention for the reasoning paradigms (Appendix~\ref{app:training}). 
We additionally report results for the baselines from~\citet{fevryEmpiricalEvaluationPretraining2020,yamadaGlobalEntityDisambiguation2022,decaoAutoregressiveEntityRetrieval2021,wangEntityDisambiguationFusion2024a,milichZELDAComprehensiveBenchmark2023}.

\paragraph{Results.} Results are reported in Table~\ref{tab:zelda_results}.
All trained configurations at 1.7B and above match or outperform FusionED, and the best 4B
configurations exceed LELA with our 26B teacher. Reasoning does not pay off in this setting: with
abundant in-domain supervision, pointwise reranking is the best configuration at 4B. Reasoning only leads at
the 0.6B scale, and on the adversarial \textit{Shadowlinks-Shadow} and \textit{-Top} splits where
mention priors are deliberately misleading; two symptoms of the same condition, reasoning paying
off only where the pointwise stage is itself unreliable (see Section~\ref{sec:analyses}).

\begin{table*}
  \centering
  \small
  \setlength{\tabcolsep}{1.0pt}
  \begin{tabular}{@{}lcccccccccc@{}}
    \toprule
    Model & AIDA & Twki & R-Post & R-Cmnt & W-CWEB & W-Wiki & S-Tail & S-Shdw & S-Top & Avg \\
    \midrule
    \multicolumn{11}{c}{\textbf{Baselines}} \\
    \midrule
    FusionED & 80.1 & 81.4 & 93.9 & 92.3 & 73.6 & 89.0 & 98.3 & 41.5 & 57.9 & 78.7 \\
    \multicolumn{11}{l}{LELA w/} \\
    \rowcolor{lightgray} \texttt{gemma-4-26B-A4B} & \ci{86.9}{1.0} & \ci{80.8}{2.6} & \ci{88.5}{2.4} & \ci{87.6}{2.6} & \ci{71.9}{0.8} & \ci{88.2}{0.8} & \ci{97.2}{1.1} & \ci{50.6}{3.3} & \ci{67.8}{3.1} & \ci{80.0}{0.7} \\
    \texttt{Qwen3-0.6B} & \ci{72.6}{1.3} & \ci{73.3}{3.0} & \ci{74.0}{3.3} & \ci{72.1}{3.5} & \ci{58.7}{0.9} & \ci{76.5}{1.0} & \ci{97.1}{1.1} & \ci{43.5}{3.2} & \ci{60.0}{3.2} & \ci{69.7}{0.8} \\
    \texttt{Qwen3-1.7B} & \ci{82.2}{1.1} & \ci{77.3}{2.8} & \ci{84.4}{2.7} & \ci{83.9}{2.9} & \ci{63.8}{0.9} & \ci{82.8}{0.9} & \ci{97.1}{1.1} & \ci{44.8}{3.2} & \ci{62.2}{3.2} & \ci{75.4}{0.8} \\
    \texttt{Qwen3-4B} & \ci{84.0}{1.1} & \ci{77.3}{2.8} & \ci{82.9}{2.8} & \ci{82.5}{3.0} & \ci{66.3}{0.9} & \ci{85.1}{0.9} & \ci{97.1}{1.1} & \ci{48.2}{3.3} & \ci{65.0}{3.1} & \ci{76.5}{0.5} \\
    \midrule
    \multicolumn{11}{c}{\textbf{Pointwise Reranking \& Direct Selection}} \\
    \midrule
    \texttt{Qwen3-Rerank-0.6B} & \done{\ci{81.6}{1.1}} & \done{\cibf{81.5}{2.6}} & \done{\cibf{95.3}{1.6}} & \done{\underline{\ci{92.6}{2.0}}} & \done{\underline{\ci{73.7}{0.8}}} & \done{\underline{\ci{87.7}{0.8}}} & \done{\cibf{97.3}{1.1}} & \done{\ci{41.6}{3.2}} & \done{\ci{56.8}{3.2}} & \done{\ci{78.7}{0.7}} \\
    \texttt{Qwen3-Rerank-4B}   & \done{\underline{\ci{86.9}{1.0}}} & \done{\cibf{82.3}{2.6}} & \done{\cibf{94.9}{1.6}} & \done{\cibf{94.7}{1.8}} & \done{\cibf{75.0}{0.8}} & \done{\cibf{90.3}{0.7}} & \done{\underline{\ci{97.2}{1.1}}} & \done{\ci{44.1}{3.2}} & \done{\underline{\ci{63.9}{3.1}}} & \done{\cibf{81.0}{0.7}} \\
    \texttt{Qwen3-0.6B} & \done{\ci{79.1}{1.2}} & \done{\ci{78.4}{2.8}} & \done{\ci{89.8}{2.3}} & \done{\ci{88.1}{2.5}} & \done{\ci{69.2}{0.9}} & \done{\ci{87.0}{0.8}} & \done{\underline{\ci{97.2}{1.1}}} & \done{\ci{41.9}{3.2}} & \done{\ci{57.2}{3.2}} & \done{\ci{76.4}{0.7}} \\
    \texttt{Qwen3-1.7B} & \done{\ci{84.2}{1.1}} & \done{\ci{80.5}{2.7}} & \done{\underline{\ci{91.6}{2.1}}} & \done{\ci{90.1}{2.3}} & \done{\ci{71.1}{0.9}} & \done{\cibf{88.7}{0.8}} & \done{\underline{\ci{97.2}{1.1}}} & \done{\underline{\ci{43.8}{3.2}}} & \done{\underline{\ci{61.0}{3.2}}} & \done{\ci{78.7}{0.7}} \\
    \texttt{Qwen3-4B}   & \done{\ci{84.7}{1.1}} & \done{\underline{\ci{81.4}{2.6}}} & \done{\ci{90.6}{2.2}} & \done{\ci{90.3}{2.3}} & \done{\ci{72.6}{0.8}} & \done{\ci{88.9}{0.8}} & \done{\underline{\ci{97.2}{1.1}}} & \done{\ci{46.2}{3.3}} & \done{\underline{\ci{63.8}{3.1}}} & \done{\ci{79.5}{0.7}} \\
    \midrule
    \multicolumn{11}{c}{\textbf{Selection (Test Time Scaling)}} \\
    \midrule
    \multicolumn{11}{c}{\textbf{RLVR}} \\
    \texttt{Qwen3-0.6B} & \done{\ci{82.3}{1.1}} & \done{\underline{\ci{80.2}{2.7}}} & \done{\ci{93.5}{1.8}} & \done{\cibf{93.3}{2.0}} & \done{\ci{72.3}{0.8}} & \done{\ci{82.5}{0.9}} & \done{\underline{\ci{97.1}{1.1}}} & \done{\ci{40.7}{3.2}} & \done{\underline{\ci{58.0}{3.2}}} & \done{\ci{77.8}{0.7}} \\
    \texttt{Qwen3-1.7B} & \done{\underline{\ci{86.4}{1.0}}} & \done{\underline{\ci{82.7}{2.5}}} & \done{\cibf{93.0}{1.9}} & \done{\cibf{94.7}{1.8}} & \done{\cibf{73.7}{0.8}} & \done{\ci{87.7}{0.8}} & \done{\underline{\ci{97.1}{1.1}}} & \done{\ci{42.8}{3.2}} & \done{\underline{\ci{59.9}{3.2}}} & \done{\cibf{79.8}{0.7}} \\
    \texttt{Qwen3-4B}   & \done{\underline{\ci{86.7}{1.0}}} & \done{\underline{\ci{81.4}{2.6}}} & \done{\ci{91.5}{2.1}} & \done{\ci{91.5}{2.2}} & \done{\ci{73.8}{0.8}} & \done{\ci{88.6}{0.8}} & \done{\cibf{97.3}{1.1}} & \done{\ci{47.9}{3.3}} & \done{\cibf{65.8}{3.1}} & \done{\ci{80.5}{0.7}} \\
    \multicolumn{11}{c}{\textbf{On-Policy Distillation}} \\
    \texttt{gemma-4-E2B-it} & \done{\ci{85.8}{1.0}} & \done{\cibf{83.0}{2.5}} & \done{\underline{\ci{92.0}{2.0}}} & \done{\ci{92.6}{2.0}} & \done{\underline{\ci{73.5}{0.8}}} & \done{\ci{87.0}{0.8}} & \done{\underline{\ci{97.2}{1.1}}} & \done{\underline{\ci{44.3}{3.2}}} & \done{\underline{\ci{60.7}{3.2}}} & \done{\underline{\ci{79.6}{0.7}}} \\
    \texttt{gemma-4-E4B-it} & \done{\underline{\ci{87.3}{1.0}}} & \done{\underline{\ci{81.9}{2.6}}} & \done{\ci{88.8}{2.3}} & \done{\ci{90.4}{2.3}} & \done{\ci{73.0}{0.8}} & \done{\ci{88.4}{0.8}} & \done{\underline{\ci{97.2}{1.1}}} & \done{\ci{48.1}{3.3}} & \done{\underline{\ci{65.3}{3.1}}} & \done{\ci{80.1}{0.7}} \\
    \multicolumn{11}{c}{\textbf{Off-Policy Distillation}} \\
    \texttt{Qwen3-0.6B} & \done{\cibf{86.1}{1.0}} & \done{\underline{\ci{80.5}{2.7}}} & \done{\ci{92.6}{1.9}} & \done{\underline{\ci{91.9}{2.1}}} & \done{\cibf{73.7}{0.8}} & \done{\cibf{87.9}{0.8}} & \done{\underline{\ci{97.2}{1.1}}} & \done{\cibf{44.3}{3.2}} & \done{\cibf{60.1}{3.2}} & \done{\cibf{79.4}{0.7}} \\
    \texttt{gemma-4-E2B-it} & \done{\cibf{86.7}{1.0}} & \done{\ci{81.2}{2.6}} & \done{\underline{\ci{91.9}{2.0}}} & \done{\ci{91.5}{2.2}} & \done{\ci{73.0}{0.8}} & \done{\ci{87.9}{0.8}} & \done{\cibf{97.3}{1.1}} & \done{\cibf{45.2}{3.3}} & \done{\cibf{61.6}{3.2}} & \done{\underline{\ci{79.6}{0.7}}} \\
    \texttt{gemma-4-E4B-it} & \done{\cibf{87.5}{1.0}} & \done{\underline{\ci{81.7}{2.6}}} & \done{\ci{89.6}{2.3}} & \done{\ci{90.1}{2.3}} & \done{\ci{73.6}{0.8}} & \done{\ci{88.6}{0.8}} & \done{\underline{\ci{97.2}{1.1}}} & \done{\cibf{49.6}{3.3}} & \done{\underline{\ci{65.7}{3.1}}} & \done{\ci{80.4}{0.7}} \\
    \bottomrule
  \end{tabular}
  \caption{ZELDA accuracy. \textbf{Bold}: best within size group; \underline{underline}: not significantly worse
  (Holm-corrected $\alpha{=}0.05$; paired McNemar per dataset, document-level paired randomization for Avg; baselines excluded).
  Additional results: Table~\ref{tab:zelda_crossfamily}, Appendix~\ref{sec:additional_results}.}
  \label{tab:zelda_results}
  \vspace{-\baselineskip}
\end{table*}

\subsection{The True Zero-Shot Setting}

\paragraph{Benchmark.} ZESHEL~\cite{logeswaranZeroShotEntityLinking2019} is a benchmark for evaluating ED methods in a zero-shot setting, built
from Wikia\footnote{\url{https://www.wikia.com/}}, a Wikipedia-like Web site for subjects such as fictional universes or sports. ZESHEL
consists of 8 Wikia domains for training, 4 for validation, and 4 for testing.
The test has a total of 10,000 mentions.  For each mention, the dataset provides 64 candidates retrieved by BM25. Entities correspond to Wikia pages, where the entity name is the title of the page and the entity description \mbox{is its content.}
The candidate retrieval has a recall of 68\% on the test set.

\paragraph{Configuration.} Following previous work \cite{logeswaranZeroShotEntityLinking2019,xuReadandSelectFrameworkZeroshot2023,liEffectiveFewShotNamed2022}, we consider the \textit{normalized accuracy} as the evaluation metric on ZESHEL, which is defined as the accuracy on the subset of mentions for which the correct entity is among the candidates. We use the provided candidates and the first three sentences of the entity page as the entity description.
We reuse the checkpoints from Section~\ref{sec:classical} and evaluate them directly on the test domains, matching the true zero-shot setting.

\paragraph{Results.} Results are reported in Table~\ref{tab:zeshel_results}
(per-category results in Appendix~\ref{app:categories}).
On ZESHEL, prompted small backbones collapse under domain shift, and training is what makes them viable at all.
The classical ordering then reverses: off-policy distillation is best in every size group, and at 4B it alone breaks away from the cluster of other variants, coming within about a point of the 26B teacher it distills from.
RLVR transfers particularly poorly at 0.6B, falling below direct selection.
Where the classical setting made reasoning optional, domain shift makes its training recipe consequential.
Additional true zero-shot results on biomedical acronym disambiguation
(GLADIS;~\citealp{chenGLADISGeneralLarge2023}) are reported in
Appendix~\ref{app:gladis}: the same ordering holds, with off-policy
distillation best in every size group.

\begin{table*}
  \centering
  \small
  \begin{tabular}{lccccc}
    \toprule
    \textbf{Method} & \textbf{F. Realms} & \textbf{Lego} & \textbf{Star Trek} & \textbf{YuGiOh} & \textbf{Macro Acc.} \\
    \midrule
    \multicolumn{6}{c}{\textbf{Baselines}} \\
    \midrule
    BLINK~\cite{wuScalableZeroshotEntity2020} & 84.11 & 74.45 & 72.22 & 66.30 & 74.27 \\
    LELA~\cite{haffoudhiLELALLMBasedEntity2026a} \\
    \rowcolor{lightgray} w/ \texttt{gemma-4-26B-A4B-it} & \ci{90.29}{1.84} & \ci{83.76}{2.32} & \ci{80.93}{1.46} & \ci{84.75}{1.56} & \ci{84.93}{0.91} \\
    w/ \texttt{Qwen3-0.6B} & \ci{10.01}{1.86} & \ci{\phantom{0}9.66}{1.86} & \ci{17.13}{1.40} & \ci{13.88}{1.50} & \ci{12.67}{0.83} \\
    w/ \texttt{Qwen3-1.7B} & \ci{36.64}{2.99} & \ci{26.21}{2.77} & \ci{35.91}{1.78} & \ci{29.57}{1.98} & \ci{32.08}{1.22} \\
    w/ \texttt{Qwen3-4B}   & \ci{80.08}{2.48} & \ci{67.73}{2.94} & \ci{65.03}{1.77} & \ci{58.26}{2.13} & \ci{67.78}{1.18} \\
    \midrule
    \multicolumn{6}{c}{\textbf{Pointwise Reranking}} \\
    \midrule
    \texttt{Qwen3-Reranker-0.6B} & \done{\ci{77.60}{2.59}} & \done{\ci{61.40}{3.06}} & \done{\ci{65.03}{1.77}} & \done{\ci{59.81}{2.12}} & \done{\ci{65.96}{1.22}} \\
    \texttt{Qwen3-Reranker-4B}   & \done{\ci{84.50}{2.25}} & \done{\ci{78.23}{2.60}} & \done{\ci{76.55}{1.57}} & \done{\ci{71.26}{1.96}} & \done{\ci{77.64}{1.06}} \\
    \midrule
    \multicolumn{6}{c}{\textbf{Direct Selection}} \\
    \midrule
    \texttt{Qwen3-0.6B} & \done{\ci{69.90}{2.85}} & \done{\ci{58.42}{3.10}} & \done{\ci{57.52}{1.84}} & \done{\ci{57.18}{2.14}} & \done{\ci{60.76}{1.27}} \\
    \texttt{Qwen3-1.7B} & \done{\ci{80.00}{2.48}} & \done{\ci{67.45}{2.95}} & \done{\ci{68.33}{1.73}} & \done{\ci{62.45}{2.10}} & \done{\ci{69.56}{1.18}} \\
    \texttt{Qwen3-4B}   & \done{\ci{83.50}{2.30}} & \done{\ci{77.52}{2.63}} & \done{\ci{73.75}{1.64}} & \done{\ci{74.53}{1.89}} & \done{\ci{77.33}{1.07}} \\
    \midrule
    \multicolumn{6}{c}{\textbf{TTS: RLVR}} \\
    \texttt{Qwen3-0.6B} & \done{\ci{56.80}{3.08}} & \done{\ci{41.68}{3.10}} & \done{\ci{49.66}{1.86}} & \done{\ci{44.86}{2.15}} & \done{\ci{48.25}{1.30}} \\
    \texttt{Qwen3-1.7B} & \done{\ci{81.80}{2.40}} & \done{\ci{70.94}{2.86}} & \done{\ci{70.77}{1.69}} & \done{\ci{66.15}{2.05}} & \done{\ci{72.42}{1.15}} \\
    \texttt{Qwen3-4B}   & \done{\ci{84.90}{2.22}} & \done{\ci{73.82}{2.77}} & \done{\ci{76.88}{1.57}} & \done{\ci{76.18}{1.84}} & \done{\ci{77.95}{1.07}} \\
    \multicolumn{6}{c}{\textbf{TTS: On-Policy Distillation}} \\
    \texttt{gemma-4-E2B-it} & \done{\ci{81.30}{2.42}} & \done{\ci{66.43}{2.97}} & \done{\ci{65.64}{1.76}} & \done{\ci{64.00}{2.08}} & \done{\ci{69.34}{1.18}} \\
    \texttt{gemma-4-E4B-it} & \done{\ci{85.90}{2.16}} & \done{\ci{74.64}{2.74}} & \done{\ci{76.19}{1.58}} & \done{\ci{74.67}{1.88}} & \done{\ci{77.85}{1.07}} \\
    \multicolumn{6}{c}{\textbf{TTS: Off-Policy Distillation}} \\
    \texttt{Qwen3-0.6B} & \done{\cibf{82.60}{2.35}} & \done{\cibf{70.33}{2.87}} & \done{\cibf{74.51}{1.62}} & \done{\cibf{70.92}{1.97}} & \done{\cibf{74.59}{1.13}} \\
    \texttt{gemma-4-E2B-it} & \done{\cibf{85.30}{2.20}} & \done{\cibf{74.44}{2.74}} & \done{\cibf{77.09}{1.56}} & \done{\cibf{76.04}{1.85}} & \done{\cibf{78.22}{1.07}} \\
    \texttt{gemma-4-E4B-it} & \done{\cibf{88.50}{1.98}} & \done{\cibf{81.52}{2.44}} & \done{\cibf{82.76}{1.40}} & \done{\cibf{81.98}{1.66}} & \done{\cibf{83.69}{0.95}} \\
    \bottomrule
  \end{tabular}
  \caption{Normalized accuracy on the test domains of ZESHEL in the true zero-shot setting. The best configuration is significantly better than every other in its group. Additional results: Table~\ref{tab:zeshel_crossfamily} (Appendix~\ref{sec:additional_results}).}
  \label{tab:zeshel_results}
\end{table*}

\subsection{The Few-Shot Setting}
\label{sec:fewshot}

\paragraph{Benchmark.} We evaluate on the four ZESHEL test domains, as for the true zero-shot setting.

\paragraph{Configuration.} For each of the four domains, we sample 100 mentions for training and test on the remaining ones, following~\citet{liEffectiveFewShotNamed2022}. 
Of the 100, we hold out 10 for validation.
We fine-tune a separate model per domain, initialized from the corresponding Wikipedia
checkpoint of Section~\ref{sec:classical}, and train for more epochs than in
the classical setting (see Appendix~\ref{app:training}) given the small mention
sets. For the TTS paradigms the initialization is the off-policy-distilled
checkpoint, following the standard two-stage LLM-reasoning training recipe~\cite{guoDeepSeekR1IncentivizesReasoning2025}. 
The listwise paradigms consume shortlists from
the unadapted ZELDA reranker (Section~\ref{sec:classical}).

\paragraph{Results.} Results are reported in Table~\ref{tab:few-shot_results}.
Limited in-domain supervision reverses the picture once more: RLVR, which transferred worst under shift, is now the best in every group. The reason is a ceiling. The distillation variants converge 
to the teacher, whereas RLVR is the one recipe to exceed a 26B model that is itself zero-shot. 
The pointwise stage recovers as well, gaining more from adaptation at 4B than any
selection paradigm.
Distilling from a 31B teacher in-domain does not improve the student (Table~\ref{tab:few-shot_additional_results}, Appendix~\ref{sec:additional_results}), suggesting teacher capacity is not the limiting factor at this scale.

\begin{table*}
  \centering
  \small
  \begin{tabular}{lccccc}
    \toprule
    \textbf{Method} & \textbf{F. Realms} & \textbf{Lego} & \textbf{Star Trek} & \textbf{YuGiOh} & \textbf{Macro Acc.} \\
    \midrule
    \multicolumn{6}{c}{\textbf{Pointwise Reranking} \textrightarrow{} \textbf{Pointwise Reranking}} \\
    \midrule
    \texttt{Qwen3-Reranker-0.6B} & \cellcolor{green!25}\done{\ci{81.87}{2.49}} & \cellcolor{green!45}\done{\underline{\ci{72.76}{2.93}}} & \cellcolor{green!30}\done{\ci{72.28}{1.68}} & \cellcolor{green!55}\done{\underline{\ci{74.79}{1.91}}} & \cellcolor{green!40}\done{\ci{75.42}{1.15}} \\
    \texttt{Qwen3-Reranker-4B} & \cellcolor{green!25}\done{\underline{\ci{86.97}{2.18}}} & \cellcolor{green!35}\done{\underline{\ci{83.86}{2.42}}} & \cellcolor{green!20}\done{\ci{79.66}{1.51}} & \cellcolor{green!60}\done{\ci{84.58}{1.59}} & \cellcolor{green!40}\done{\ci{83.77}{0.98}} \\
    \midrule
    \multicolumn{6}{c}{\textbf{Direct Selection} \textrightarrow{} \textbf{Direct Selection}} \\
    \midrule
    \texttt{Qwen3-0.6B} & \cellcolor{green!60}\done{\ci{82.74}{2.45}} & \cellcolor{green!45}\done{\ci{70.63}{2.99}} & \cellcolor{green!40}\done{\ci{69.31}{1.73}} & \cellcolor{green!60}\done{\underline{\ci{73.88}{1.93}}} & \cellcolor{green!50}\done{\ci{74.14}{1.16}} \\
    \texttt{Qwen3-1.7B} & \cellcolor{green!30}\done{\underline{\ci{84.15}{2.36}}} & \cellcolor{green!30}\done{\ci{73.32}{2.91}} & \cellcolor{green!30}\done{\ci{74.08}{1.65}} & \cellcolor{green!50}\done{\ci{74.59}{1.91}} & \cellcolor{green!35}\done{\ci{76.53}{1.13}} \\
    \texttt{Qwen3-4B} & \cellcolor{green!20}\done{\ci{85.67}{2.27}} & \cellcolor{green!35}\done{\underline{\ci{82.40}{2.50}}} & \cellcolor{green!20}\done{\ci{77.57}{1.57}} & \cellcolor{green!40}\done{\ci{81.52}{1.71}} & \cellcolor{green!30}\done{\ci{81.79}{1.02}} \\
    \midrule
    \multicolumn{6}{c}{\textbf{Selection (Test Time Scaling)}} \\
    \midrule
    \multicolumn{6}{c}{\textbf{Off-Policy Distillation} \textrightarrow{} \textbf{RLVR}} \\
    \texttt{Qwen3-0.6B} & \cellcolor{green!30}\done{\cibf{86.21}{2.23}} & \cellcolor{green!20}\done{\cibf{74.55}{2.86}} & \done{\underline{\ci{74.52}{1.64}}} & \cellcolor{green!20}\done{\cibf{75.24}{1.90}} & \cellcolor{green!15}\done{\cibf{77.63}{1.10}} \\
    \texttt{gemma-4-E2B-it} & \cellcolor{red!10}\done{\ci{84.47}{2.34}} & \cellcolor{green!25}\done{\cibf{79.48}{2.66}} & \cellcolor{green!20}\done{\cibf{79.77}{1.51}} & \cellcolor{green!40}\done{\cibf{82.87}{1.66}} & \cellcolor{green!25}\done{\cibf{81.65}{1.05}} \\
    \texttt{gemma-4-E4B-it} & \done{\underline{\ci{88.27}{2.08}}} & \cellcolor{green!25}\done{\cibf{84.19}{2.40}} & \cellcolor{green!5}\done{\cibf{83.48}{1.40}} & \cellcolor{green!45}\done{\cibf{87.74}{1.44}} & \cellcolor{green!20}\done{\cibf{85.92}{0.94}} \\
    \midrule
    \multicolumn{6}{c}{\textbf{Off-Policy Distillation} \textrightarrow{} \textbf{On-Policy Distillation}} \\
    \texttt{gemma-4-E2B-it} & \cellcolor{green!10}\done{\underline{\ci{86.21}{2.23}}} & \cellcolor{red!5}\done{\ci{74.33}{2.87}} & \cellcolor{red!10}\done{\ci{75.29}{1.62}} & \cellcolor{green!15}\done{\ci{78.35}{1.81}} & \done{\ci{78.55}{1.09}} \\
    \texttt{gemma-4-E4B-it} & \done{\underline{\ci{88.60}{2.06}}} & \cellcolor{green!5}\done{\underline{\ci{81.95}{2.53}}} & \done{\ci{82.45}{1.43}} & \cellcolor{green!15}\done{\ci{83.88}{1.62}} & \cellcolor{green!5}\done{\ci{84.22}{0.98}} \\
    \multicolumn{6}{c}{\textbf{Off-Policy Distillation} \textrightarrow{} \textbf{Off-Policy Distillation}} \\
    \texttt{Qwen3-0.6B} & \cellcolor{green!10}\done{\ci{83.82}{2.38}} & \cellcolor{green!15}\done{\underline{\ci{72.98}{2.92}}} & \done{\cibf{74.89}{1.63}} & \done{\ci{71.22}{1.99}} & \cellcolor{green!5}\done{\ci{75.73}{1.14}} \\
    \texttt{gemma-4-E2B-it} & \cellcolor{green!10}\done{\cibf{86.54}{2.21}} & \cellcolor{green!20}\done{\underline{\ci{78.14}{2.72}}} & \cellcolor{green!15}\done{\underline{\ci{79.15}{1.53}}} & \cellcolor{green!35}\done{\underline{\ci{81.72}{1.70}}} & \cellcolor{green!20}\done{\underline{\ci{81.39}{1.05}}} \\
    \texttt{gemma-4-E4B-it} & \cellcolor{green!5}\done{\cibf{88.82}{2.04}} & \cellcolor{green!15}\done{\underline{\ci{83.30}{2.45}}} & \done{\underline{\ci{82.67}{1.42}}} & \cellcolor{green!20}\done{\ci{84.83}{1.58}} & \cellcolor{green!10}\done{\ci{84.90}{0.96}} \\
    \bottomrule
  \end{tabular}
  \caption{Normalized accuracy on the test domains of ZESHEL in the few-shot setting.
  Arrows denote the classical-setting checkpoint used as initialization followed by the few-shot training paradigm.
  Shading gives the relative error reduction over that checkpoint, $\Delta/(100-\text{base})$.
  Additional results: Table~\ref{tab:few-shot_additional_results} (Appendix~\ref{sec:additional_results}).}
  \label{tab:few-shot_results}
\end{table*}
 
\subsection{Analyses}
\label{sec:analyses}

\begin{figure}[t]
\centering
\includegraphics[width=\columnwidth]{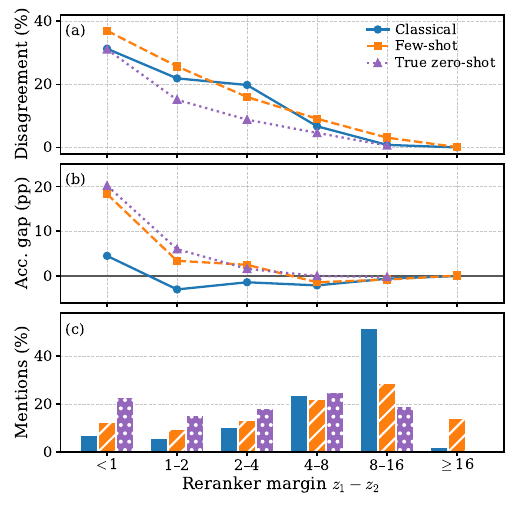}
\caption{(a) Percentage of mentions where exactly one system is correct, (b) accuracy difference (reasoning $-$ reranking, in percentage points), and (c) mention distribution, by reranker margin $z_1-z_2$.}
\label{fig:margin_mediation}
\end{figure}

\paragraph{Uncertainty locates disagreement.}
Reranking and reasoning fail on much the same mentions, but which
system is right when they disagree changes across settings
(Table~\ref{tab:complementarity-main}). In the classical setting,
reasoning breaks slightly more correct reranker predictions than
it repairs; under true zero-shot shift, it repairs three
times as many as it breaks.
Writing $z_1-z_2$ for the log-odds gap between the reranker's top
two candidates, disagreement falls steeply as this margin grows,
while low-margin mentions become more frequent under shift
(Figure~\ref{fig:margin_mediation}a,c).

\paragraph{Domain shift favors reasoning.}
To distinguish changes in uncertainty from changes in which
paradigm succeeds, we reweight each setting's margin bins to the
classical margin distribution. This reduces true-zero-shot
disagreement from $12.40\%$ to $5.40\%$, but reasoning still repairs
about twice as many predictions as it breaks. Thus, the greater
frequency of uncertain mentions largely accounts for the increased
disagreement, but does not fully explain its direction
(Figure~\ref{fig:margin_mediation}b).
The margin also discriminates reranker correctness more strongly
than it identifies the correct system when exactly one succeeds:
$\mathrm{AUC}_{\mathrm{corr}}$ ranges from $0.78$ to $0.85$,
versus $0.61$ to $0.68$ for $\mathrm{AUC}_{\mathrm{route}}$
(Table~\ref{tab:complementarity-main}).
Consistent with this limited routing signal, even test-optimized
margin routing gains at most $1.1$ accuracy points over the better
standalone method, and only $0.29$ under true zero-shot shift.
Standardization and routing details are given in
Appendix~\ref{app:complementarity}.

\begin{table}
\centering
\small
\setlength{\tabcolsep}{4pt}
\begin{tabular}{lcccc}
\toprule
Setting & Repairs & Breaks
  & $\mathrm{AUC}_{\mathrm{corr}}$
  & $\mathrm{AUC}_{\mathrm{route}}$ \\
\midrule
Classical      & 2.9 & 3.0 & 0.85 & 0.61 \\
Few-shot       & 6.6 & 4.5 & 0.83 & 0.68 \\
True zero-shot & 9.2 & 3.2 & 0.78 & 0.63 \\
\bottomrule
\end{tabular}
\caption{Complementarity and margin discrimination at 4B,
using the best reasoning variant per setting.
Repairs (\%): reranker wrong, reasoning correct;
breaks (\%): reranker correct, reasoning wrong.
$\mathrm{AUC}_{\mathrm{corr}}$: margin-based prediction
of reranker correctness; $\mathrm{AUC}_{\mathrm{route}}$:
prediction of reranker correctness conditional on exactly one
system being correct (Appendix~\ref{app:complementarity}).}
\label{tab:complementarity-main}
\vspace{-\baselineskip}
\end{table}

\paragraph{Gains concentrate in lexical ambiguity.}
At 4B on ZESHEL, off-policy distillation improves over pointwise
reranking by 8.7 percentage points on \textit{Multiple Categories} mentions,
compared with 4.0 on \textit{Low Overlap} and at most 1.2 on the other
categories (Appendix~\ref{app:categories}). \textit{Multiple Categories} mentions match the entity title apart from a disambiguation phrase, suggesting that
reasoning is particularly useful when surface-form matching alone
leaves the referent ambiguous. It is also the category where scaling the reranker buys the least.

\begin{figure}[t]
    \centering
    \includegraphics[width=\linewidth]{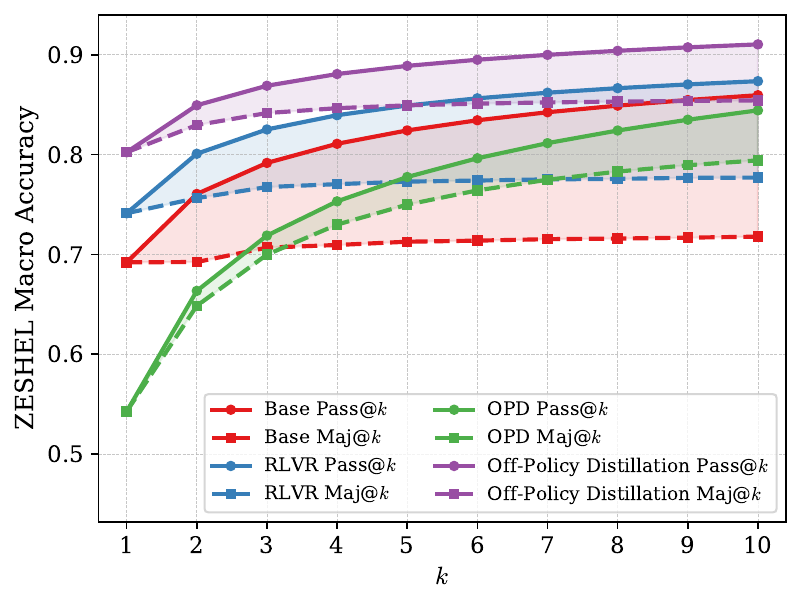}
    \caption{Pass@$k$ and Maj@$k$ on ZESHEL (true zero-shot), 4B. All sizes: Figure~\ref{fig:passk}, Appendix~\ref{app:tts}}
    \label{fig:passmaj-main}
    \vspace{-\baselineskip}
\end{figure}

\paragraph{Equal accuracy, unequal cost.}
Beyond accuracy, the training recipes lead to different output distributions
(Figure~\ref{fig:passmaj-main}). Under RLVR, majority voting recovers no more
than a single sample: correct answers do appear among the samples (Pass@$k$
rises with $k$), but rarely as the majority, so self-consistency adds nothing
(Appendix~\ref{app:tts}). Distillation is the opposite: its votes track its
Pass@$k$, which under shift voting converts into accuracy, but it pays for this
with far longer traces that lengthen as training proceeds, in all three settings, and at 2B degrade on-policy accuracy (Appendix~\ref{app:training_dynamics}).

\paragraph{Adaptation at the cost of forgetting.}
Domain adaptation comes with a caveat that per-domain accuracy
cannot show. Adapting with in-domain RLVR degrades source-domain
accuracy on AIDA by 2.0 to 5.8 percentage points, whereas both
distillation variants limit the loss to at most 1.4 points
(Table~\ref{tab:continual_results}, Appendix~\ref{app:continual}).
The recipe that best exploits target supervision is thus the one that specializes
into it. Where a single model must
serve several domains, distillation is the safer adaptation step even
where RLVR scores higher on the target.

\paragraph{Does on-policy training explain the gap?}
Our main distillation recipes differ in both trajectory source
and KL direction. To distinguish their effects, we cross teacher-
and student-generated trajectories with forward and reverse KL,
holding the 4B student and teacher fixed
(Table~\ref{tab:trajectory-kl}).
On true-zero-shot ZESHEL, student and teacher trajectories
achieve nearly identical normalized accuracy under forward KL.
Under reverse KL, however, student trajectories perform worse
than teacher trajectories.
Forward KL therefore improves accuracy with either trajectory
source, with a substantially larger gain on student trajectories.
These results indicate that the effect of on-policiness depends
on the distillation objective: the advantage of our standard
off-policy recipe cannot be attributed to teacher-generated
trajectories alone.

\begin{table}
    \centering
    \small
    \begin{tabular}{lcc}
        \toprule
        Trajectory source
            & Forward KL & Reverse KL \\
            & ($\beta=0$) & ($\beta=1$) \\
        \midrule
        Teacher ($\lambda=0$) & 83.69 & 82.68 \\
        Student ($\lambda=1$) & 83.76 & 77.85 \\
        \bottomrule
    \end{tabular}
    \caption{Macro normalized accuracy on true-zero-shot
    ZESHEL, crossing trajectory source and KL direction.}
    \label{tab:trajectory-kl}
    \vspace{-\baselineskip}
\end{table}

\subsection{Discussion}
No single paradigm dominates, and the setting taxonomy alone does not say
which one to use. Reranker uncertainty is a second, partly independent axis:
it locates the mentions where the paradigms diverge (examples in
Appendix~\ref{app:examples}), while the setting decides which of them is
right there. That uncertainty has a lexical counterpart in ZESHEL's mention
categories, and complements the established setting
taxonomy and entity-frequency strata~\cite{decaoAutoregressiveEntityRetrieval2021}
as an axis along which to compare ED systems.

For training, off-policy distillation transfers better than RLVR under shift, whereas RLVR leads in few-shot adaptation by specializing to the target domain.
On ED at this scale, the familiar SFT-memorizes / RL-generalizes reading~\citep{chuSFTMemorizesRL2025a} runs the wrong way.

\section{Conclusion}

We studied small LLM backbones for
entity disambiguation, comparing candidate-selection paradigms and
training recipes for reasoning-based selection across the
classical, few-shot and true zero-shot settings. The resulting mapping is
concrete: pointwise reranking suffices in the classical setting, where
reasoning does not repay its inference cost; RLVR from a distilled
base is best given even a hundred target-domain mentions; and
off-policy distillation performs best among the three main training
recipes under pure domain shift,
bringing a 4B model within 1.2 points of its 26B teacher. 
ED reasoning is therefore not the preserve of large models. 
The mapping is not governed by the shift axis alone: reasoning pays where the
reranker is uncertain over its candidates, and shift matters both because it
drives that uncertainty up and because it decides which paradigm wins there.
Our code is available on GitHub\footnote{\url{https://github.com/samyhaff/ed-study}}
and our models on Hugging Face\footnote{\url{https://hf.co/collections/samyhaff/ed-study}}.

\section*{Limitations}

\paragraph{Scale and model coverage.} Our study spans 0.6--4B backbones from
two model families (Qwen3 and Gemma-4), distilled from 26--31B teachers. The
recommendations we draw are specific to this range: we cannot say whether the
paradigm ordering we observe holds for 8B+ students, different LLM-based rerankers,
or whether it is stable across architectures we did not test. 
Nonetheless, our scale selection remains aligned with the typical sizes of rerankers,
with larger model sizes being mostly limited to generative paradigms.

\paragraph{Benchmarks and shift.} We evaluate on English only, with domain shift instantiated by ZESHEL’s Wikia domains. While evaluation was conducted on four distinct domains, these are fictional universes with small, self-contained KBs, which is one particular kind of shift; whether our true zero-shot conclusions transfer to shifts of a different character remains open. Results on biomedical acronym disambiguation, which shifts both domain and task, suggest that the paradigm ordering does.

\paragraph{Coverage of the training design space.} Our main comparison covers two GKD endpoints, supplemented by a $2 \times 2$ ablation; intermediate settings remain unexplored. RLVR is represented by GRPO alone, and all runs use LoRA rather than full
fine-tuning. Hyperparameters were held fixed across paradigms for
comparability rather than tuned per paradigm, which may not favor each
equally.

\paragraph{Statistical scope.} Each configuration corresponds to a single training run, so variance across seeds and data orderings is not quantified. The
claims we draw, however, are orderings that reproduce across three model sizes, two model families, and independent domains, rather than differences between individual runs; the per-domain results on ZESHEL, the results on GLADIS and the cross-family replication each constitute an independent instantiation. Multi-seed replication of the full grid would multiply the training compute several-fold.

\FloatBarrier

\section*{Acknowledgements}

The work was partially supported by Agence de l’Innovation de Défense AID - via Centre Interdisciplinaire d’Etudes pour la Défense et la Sécurité CIEDS - (project 2024 - KB-LM). Furthermore, we thank GENCI at IDRIS  for providing access to computational resources through the grant 2026-AD011017960 on the supercomputer Jean Zay.

\bibliography{lela2}

\FloatBarrier

\appendix

\begin{table}[t]
\centering
\small
\renewcommand{\arraystretch}{1.2}
\setlength{\tabcolsep}{3pt}
\begin{tabularx}{\columnwidth}{@{}l CCC@{}}
\toprule
 & \textbf{Pointwise} & \textbf{Direct} & \textbf{TTS} \\
 & \textbf{Reranking} & \textbf{Selection} & \textbf{Selection} \\
\midrule
Formulation        & Binary       & Listwise & Listwise \\
Recall ceiling     & cand.\ gen.\ & Acc@$k$ & Acc@$k$ \\
Sees all cand.\         & \cmark & \xmark & \xmark \\
Cross-cand.\ comp.\     & \xmark & \cmark & \cmark \\
Real-valued score       & \cmark & \xmark & \xmark \\
Reasoning trace         & \xmark & \xmark & \cmark \\
\addlinespace[2pt]
LLM passes / ment.\  & $n_c$ & $1$ & $n$ \\
Training signal      & BCE & CE (SFT) & RL / dist. \\
\bottomrule
\end{tabularx}
\caption{Comparison of the three candidate-selection paradigms. $n_c=|C(m,E)|$,
$k$ is the shortlist size, $n$ the number of sampled completions.}
\label{tab:paradigm-comparison}
\end{table}

\section{Prompts}
\label{sec:prompts}

The reranker instruction is given in Figure~\ref{fig:reranker_prompt}, and the selection prompt in Figure~\ref{fig:selection_prompt}. Braces denote fields instantiated per mention.

\begin{figure}[t]
\centering

\begin{subfigure}{\columnwidth}
\begin{promptbox}
\ttfamily\footnotesize\raggedright
Given a text with a marked mention enclosed in square brackets, retrieve
relevant entities that the mention refers to.
\end{promptbox}
\caption{Reranker task instruction. The query is the context with the
mention marked, and the document the candidate's title and description.}
\label{fig:reranker_prompt}
\end{subfigure}

\vspace{0.6\baselineskip}

\begin{subfigure}{\columnwidth}
\begin{promptbox}[title=\footnotesize\bfseries System]
\ttfamily\footnotesize\raggedright
You are an expert designed to disambiguate entities in text, taking into
account the overall context and a list of entity candidates. You are
provided with an input text that includes a full contextual narrative, a
marked mention enclosed in square brackets, and a list of candidates, each
preceded by an index number.\\[0.6ex]
Your task is to determine the most appropriate entity from the candidates
based on the context and candidate entity descriptions. Please show your
choice with only the index, e.g., ``answer: 3''.
\end{promptbox}
\begin{promptbox}[title=\footnotesize\bfseries User]
\ttfamily\footnotesize\raggedright
Text: \{context with [mention] marked\}\\[0.6ex]
Candidates:\\
1. \{name\} -- \{description\}\\
2. \{name\} -- \{description\}\\
\dots\\
\{k\}. \{name\} -- \{description\}
\end{promptbox}
\caption{Selection prompt, shared by direct selection and the
test-time-scaling paradigms.}
\label{fig:selection_prompt}
\end{subfigure}

\caption{Prompts used in our pipeline.}
\label{fig:prompts}
\end{figure}

\section{Training Objectives}
\label{app:objectives}

\begin{table}
\centering
\small
\renewcommand{\arraystretch}{1.2}
\setlength{\tabcolsep}{3pt}
\begin{tabularx}{\columnwidth}{@{}l CCC@{}}
\toprule
 & \textbf{Off-policy} & \textbf{On-policy /} & \textbf{RLVR} \\
 & \textbf{Distill.} & \textbf{Distill.} & \textbf{(GRPO)} \\
\midrule
GKD $(\lambda,\beta)$ & $(0,0)$ & $(1,1)$ & --- \\
\addlinespace[2pt]
Trajectories       & teacher & student & student \\
Resampled          & \xmark & \cmark & \cmark \\
Divergence         & fwd.\ KL & rev.\ KL & --- \\
Objective          & CE & KL & PG \\
\addlinespace[2pt]
Signal granul.\    & token & token & sequence \\
Dense signal       & \cmark & \cmark & \xmark \\
Gold label         & filter & filter & reward \\
Teacher at train   & offline/online & online & \xmark \\
\addlinespace[2pt]
Mention filter     & teach.\ corr.\ & teach.\ corr. & pass band \\
Acc.\ ceiling      & teach.\ traces & teach.\ policy & base supp.\ \\
\bottomrule
\end{tabularx}
\caption{Comparison of the three training paradigms for
reasoning-based selection. $(\lambda,\beta)$ are the coefficients of
Eq.~\ref{eq:gkd}; PG denotes the clipped policy-gradient objective of
Appendix~\ref{app:grpo}. \emph{Gold label}: distillation uses $e^*$ only to
filter mentions (Section~\ref{sec:data}), while RLVR uses it as the reward. \emph{Resampled}: whether trajectories are
regenerated from the current policy at every pass rather than precomputed,
which matters when the mention set is small. \emph{Acc.\ ceiling}: the
teacher for distillation, the base model's own sampling support for RLVR.}
\label{tab:training-comparison}
\vspace{-\baselineskip}
\end{table}

We give the explicit form of the objectives summarised in
Section~\ref{sec:training} (Eq.~\ref{eq:gkd}). Throughout,
$x = (T, m, C^{(k)}_m)$ denotes the prompt, and $p_T(\cdot \mid x)$
and $p_\theta(\cdot \mid x)$ the teacher and student next-token
distributions. All divergences are computed per token position and
averaged over the completion.

\subsection{Divergences}
\label{app:divergences}

The forward and reverse KL divergences between teacher and student
are
\begin{align}
  D_{\mathrm{KL}}(p_T \Vert p_\theta)
    &= \mathbb{E}_{y \sim p_T}\!\left[ \log \frac{p_T(y)}{p_\theta(y)} \right],
    \label{eq:fkl}\\[2pt]
  D_{\mathrm{KL}}(p_\theta \Vert p_T)
    &= \mathbb{E}_{y \sim p_\theta}\!\left[ \log \frac{p_\theta(y)}{p_T(y)} \right],
    \label{eq:rkl}
\end{align}
the former mean-seeking (it penalises the student for placing no mass
where the teacher does) and the latter mode-seeking (it penalises the
student for placing mass where the teacher does not).

The generalized Jensen--Shannon divergence \citep{englessonGeneralizedJensenshannonDivergence2021a}
interpolates between the two through the mixture
$p_\beta = \beta\, p_T + (1-\beta)\, p_\theta$:
\begin{equation}
  \begin{split}
    D^{\beta}_{\mathrm{JSD}}(p_T \Vert p_\theta)
      = {}& \beta\, D_{\mathrm{KL}}(p_T \Vert p_\beta) \\
      &+ (1-\beta)\, D_{\mathrm{KL}}(p_\theta \Vert p_\beta).
  \end{split}
  \label{eq:jsd}
\end{equation}
The rescaled divergence recovers the two KLs:
\begin{equation}
  \begin{split}
    \lim_{\beta \to 0} \tfrac{1}{\beta} D^{\beta}_{\mathrm{JSD}}
      &= D_{\mathrm{KL}}(p_T \Vert p_\theta), \\
    \lim_{\beta \to 1} \tfrac{1}{1-\beta} D^{\beta}_{\mathrm{JSD}}
      &= D_{\mathrm{KL}}(p_\theta \Vert p_T).
  \end{split}
  \label{eq:jsd-limits}
\end{equation}
As a constant rescaling of the loss is absorbed into the
learning rate, we refer to $\beta = 0$ and $\beta = 1$ as the forward-
and reverse-KL settings.

\subsection{Distillation regimes}
\label{app:gkd}

The two terms of Eq.~\ref{eq:gkd} differ only in the distribution over
which the divergence is evaluated. Off-policy distillation uses
teacher-generated completions $(x, y) \in \mathcal{D}_{\mathrm{distill}}$,
\begin{equation}
  \mathcal{L}_{\mathrm{OFF}}(\theta)
    = \mathbb{E}_{(x,y) \sim \mathcal{D}_{\mathrm{distill}}}
      \big[ D^{\beta}_{\mathrm{JSD}}(p_T \Vert p_\theta)(y \mid x) \big],
  \label{eq:sd}
\end{equation}
while on-policy distillation uses completions sampled from the current
student,
\begin{equation}
  \mathcal{L}_{\mathrm{ON}}(\theta)
    = \mathbb{E}_{x \sim \mathcal{X}}\,
      \mathbb{E}_{\hat{y} \sim p_\theta(\cdot \mid x)}
      \big[ D^{\beta}_{\mathrm{JSD}}(p_T \Vert p_\theta)(\hat{y} \mid x) \big],
  \label{eq:od}
\end{equation}
and $\mathcal{L}_{\mathrm{GKD}}(\theta, \lambda, \beta) =
(1-\lambda)\mathcal{L}_{\mathrm{OFF}} + \lambda \mathcal{L}_{\mathrm{ON}}$
recovers Eq.~\ref{eq:gkd}. We study the two endpoints. At
$(\lambda, \beta) = (0,0)$ the student is trained on the teacher's
correct traces under the forward KL; retaining the teacher's sampled
tokens as hard targets collapses Eq.~\ref{eq:fkl} to the token-level
cross-entropy $-\mathbb{E}[\log p_\theta(y \mid x)]$, i.e.\ SFT on those
completions. At $(\lambda, \beta) = (1,1)$ trajectories are drawn from
the student and scored online by the teacher under the reverse KL.

\paragraph{Reduced loss support.} Computing $D^{\beta}_{\mathrm{JSD}}$ over the
full vocabulary requires transferring the teacher's complete logit tensor at
every position. We instead use the top-$k$ approximation of \texttt{trl}'s \texttt{DistillationTrainer} with 
$k=1$: both distributions are restricted to a single token per position plus a tail bucket collecting the residual probability mass, then renormalized. The retained token follows the direction of the divergence: the teacher's top-1 token for the forward KL, and the sampled completion token for the reverse KL.

\paragraph{Policy-gradient view.} On-policy distillation is equivalent
to a policy-gradient update whose token-level advantage is the
teacher--student log-likelihood ratio \citep{guMiniLLMKnowledgeDistillation2023}. 
For a student completion $\hat{y}$, let
\begin{equation}
  A_n(x, \hat{y}) = \log
    \frac{p_T(\hat{y}_n \mid x, \hat{y}_{<n})}
         {p_\theta(\hat{y}_n \mid x, \hat{y}_{<n})}.
  \label{eq:tokadv}
\end{equation}
The gradient of Eq.~\ref{eq:od} then takes the REINFORCE form
\begin{multline}
  \nabla_\theta \mathcal{L}_{\mathrm{ON}}(\theta) = -\, \mathbb{E}_{x}\,
    \mathbb{E}_{\hat{y} \sim p_\theta(\cdot \mid x)} \\
  \left[ \frac{1}{|\hat{y}|} \sum_{n=1}^{|\hat{y}|}
    A_n \nabla_\theta \log p_\theta(\hat{y}_n \mid x, \hat{y}_{<n}) \right]
\end{multline}
the gradient flowing only through the student log-probability
($A_n$ is treated as a stop-gradient quantity). On-policy distillation
and RLVR are therefore both policy-gradient methods on
student-sampled trajectories, differing only in the advantage signal:
a dense per-token teacher log-ratio in one case, a single binary
task reward in the other.

\subsection{GRPO objective}
\label{app:grpo}

For each prompt we sample $G$ completions $\{y^{(g)}\}_{g=1}^{G}$ from
$\pi_\theta$ and form group-normalised advantages
$A^{(g)} = (R(y^{(g)}) - \mu)/\sigma$, with $\mu, \sigma$ the
within-group reward mean and standard deviation. Writing
$\rho^{(g)}_t = \pi_\theta(y^{(g)}_t \mid y^{(g)}_{<t}, x) /
\pi_{\theta_{\mathrm{old}}}(y^{(g)}_t \mid y^{(g)}_{<t}, x)$ for the
importance ratio, the objective is
\begin{multline}
  \mathcal{J}_{\mathrm{GRPO}}(\theta) = \mathbb{E}
    \Bigg[ \frac{1}{G} \sum_{g=1}^{G} \frac{1}{|y^{(g)}|}
      \sum_{t=1}^{|y^{(g)}|} \\
    \min\!\Big( \rho^{(g)}_t A^{(g)},\,
      \mathrm{clip}\big(\rho^{(g)}_t, 1-\epsilon, 1+\epsilon\big) A^{(g)} \Big)
    \Bigg] \\
    - \beta_{\mathrm{KL}}\, D_{\mathrm{KL}}\big[\pi_\theta \Vert \pi_{\mathrm{ref}}\big],
  \label{eq:grpo}
\end{multline}
where $\epsilon$ is the clipping parameter and $\pi_{\mathrm{ref}}$ the
frozen initial policy.
$\beta_{\mathrm{KL}} = 0$, i.e.\ no regularisation towards the
reference policy.

\section{Training Data and Hyperparameters}
\label{app:training}

\paragraph{Training pool (classical setting).} From the ZELDA Wikipedia training split we sample 55{,}000 documents at random, taking one mention per document to favour document diversity, and keeping only documents under 3{,}072 words and mentions with at least 10 candidates. We hold out 5{,}000 mentions for validation and checkpoint selection.

\paragraph{Hyperparameters.}
All runs use LoRA ($r{=}128$, $\alpha{=}256$, all linear layers), AdamW with a
linear schedule and no warmup, and an effective batch size of 16. Learning rate
depends on model size and paradigm. Table~\ref{tab:hyperparams} lists the
per-paradigm settings.

\begin{table*}
\centering
\small
\setlength{\tabcolsep}{4pt}
\begin{tabular}{@{}l c c c l l@{}}
\toprule
Paradigm & \multicolumn{2}{c}{Max Epochs} & Max Len. & Learning Rate & Other \\
\cmidrule(lr){2-3}
 & Class. & Few-shot & & & \\
\midrule
Reranking          & 1 & 15 & 4096 & $1{\times}10^{-5}$ (all) & --- \\
Direct sel.\ (SFT) & 3 & 100 & 5120 & $1{\times}10^{-4}$ (0.6B), $1{\times}10^{-5}$ ($\ge$1.7B) & --- \\
RLVR               & 3 & 15 & 8192 & $1{\times}10^{-5}$ (all) & $G{=}16$, temp.\ 1.0, $\epsilon = 0.2$, $\beta_{\mathrm{KL}} = 0$ \\
Off-policy dist.\  & 3 & 200 & 8192 & $1{\times}10^{-4}$ (0.6B), $1{\times}10^{-5}$ ($\ge$1.7B) & $(\lambda, \beta) = (0,0)$ \\
On-policy dist.\   & 3 & 200 & 8192 & $1{\times}10^{-4}$ (0.6B), $1{\times}10^{-5}$ ($\ge$1.7B) & $(\lambda, \beta) = (1,1)$, student's gen.\ config. \\
\bottomrule
\end{tabular}
\caption{Training hyperparameters for the classical and few-shot settings. All other hyperparameters are shared.}
\label{tab:hyperparams}
\end{table*}

\section{Compute}
\label{app:compute}

\subsection{Training}
\label{app:training-cost}

Table~\ref{tab:training-cost} reports the compute consumed by models we trained on ZELDA, in
floating-point operations and in training tokens.

Following standard practice \citep{kaplanScalingLawsNeural2020}, we estimate a forward pass over $D$
tokens at $2ND$ FLOPs and a forward--backward pass at $6ND$, where $N$ is the number of
non-embedding parameters, LoRA adapters included; for mixture-of-experts teachers we use
active rather than total parameters. The total cost of a run aggregates every pass it
performs,
\begin{equation}
C = \underbrace{6 N_{s} D}_{\text{training}}
  + \underbrace{2 N_{t} D}_{\text{teacher}}
  + \underbrace{2 N_{s} D_{\text{gen}}}_{\text{generation}} ,
\end{equation}
where $N_s$ and $N_t$ denote student and teacher size. The teacher term applies only to
distillation, where a frozen teacher scores every training sequence, and the generation term
only where a model produces its own completions, i.e.\ on-policy distillation and GRPO
rollouts. No reference-model or old-policy passes appear, as we train GRPO with $\beta_\text{KL} = 0$.

$D$ is the number of tokens processed by the training pass, summed over the run: a token seen in two epochs is counted twice, and prompt tokens are counted even where the loss is restricted to the completion, since they are processed regardless. Under GRPO each prompt yields $G$ completions, and its prompt tokens are therefore counted $G$ times. Tokens processed while generating rollouts are counted in $D_{\text{gen}}$, which contributes to $C$ but not to $D$: a prompt is prefilled once per prompt, yet trained on $G$ times.

\begin{table}[t]
\centering
\small
\setlength{\tabcolsep}{3.0pt}
\begin{tabular}{llrr}
\toprule
Model & Paradigm & FLOPs & Tokens \\
      &          & ($\times 10^{18}$) & (M) \\
\midrule
\texttt{Qwen3-0.6B}     & Direct selection & 0.25 & 81 \\
\texttt{Qwen3-1.7B}     & Direct selection & 0.75 & 81 \\
\texttt{Qwen3-4B}       & Direct selection & 3.79 & 162 \\
\midrule
\texttt{Qwen3-0.6B}     & Reranking & 0.48 & 153 \\
\texttt{Qwen3-4B}       & Reranking & 2.48 & 106 \\
\midrule
\texttt{Qwen3-0.6B}     & Off-policy distill. & 0.67 & 78 \\
\texttt{Qwen3-1.7B}     & Off-policy distill. & 0.79 & 53 \\
\texttt{Qwen3-4B}       & Off-policy distill. & 1.27 & 44 \\
\texttt{gemma-4-E2B-it} & Off-policy distill. & 1.59 & 72 \\
\texttt{gemma-4-E4B-it} & Off-policy distill. & 2.94 & 83 \\
\midrule
\texttt{Qwen3-0.6B}     & On-policy distill. & 0.79 & 82 \\
\texttt{Qwen3-1.7B}     & On-policy distill. & 0.90 & 50 \\
\texttt{Qwen3-4B}       & On-policy distill. & 4.60 & 125 \\
\texttt{gemma-4-E2B-it} & On-policy distill. & 0.85 & 31 \\
\texttt{gemma-4-E4B-it} & On-policy distill. & 5.14 & 114 \\
\midrule
\texttt{Qwen3-0.6B}     & GRPO & 0.15 & 42 \\
\texttt{Qwen3-1.7B}     & GRPO & 5.53 & 526 \\
\texttt{Qwen3-4B}       & GRPO & 4.89 & 183 \\
\texttt{gemma-4-E2B-it} & GRPO & 6.71 & 358 \\
\texttt{gemma-4-E4B-it} & GRPO & 8.00 & 235 \\
\bottomrule
\end{tabular}
\caption{Training compute per model and paradigm for the chosen checkpoints. FLOPs include the frozen teacher
(\texttt{gemma-4-26B-A4B-it} or \texttt{Qwen3-30B-A3B})
and all rollout generation. Tokens is the number of tokens seen by the training pass over the whole run.}
\label{tab:training-cost}
\vspace{-\baselineskip}
\end{table}

\subsection{Inference}
\label{app:inference-cost}

Figure~\ref{fig:results_summary} plots accuracy against the inference cost of linking a single
mention, in floating-point operations. We keep the 2 FLOPs per parameter per token
convention of Appendix~\ref{app:training-cost}, applied to both prefill and decoding, and
count the two stages every generative paradigm shares,
\begin{equation}
C = \underbrace{2 N_{r} T_{r}}_{\text{candidate prefilter}}
  + \underbrace{2 N_{s} (T_{\text{in}} + T_{\text{out}})}_{\text{disambiguation}} ,
\end{equation}
where $N_r$ and $N_s$ are the non-embedding parameter counts of the reranker and of the
disambiguation model. Retrieval is excluded, being identical across paradigms; the prefilter
is not, since direct selection and every test-time-scaling variant consume the
top-$10$ shortlist and therefore pay for a full reranker pass before generating anything. The
pointwise reranking rows are that pass alone ($T_{\text{in}} = T_{\text{out}} = 0$). Models up
to 2B are prefiltered by the 0.6B reranker and the 4B models by the 4B reranker, matching how
each pipeline is actually run.

$T_r$ is the number of tokens the reranker prefills per mention. Scoring $K$ candidates
issues $K$ sequences that share one query prefix and differ only in the candidate document,
so under vLLM's automatic prefix caching the query is prefilled once and $T_r = T_{\text{query}}
+ \sum_{k} T_{\text{doc}_k}$. $T_{\text{in}}$
and $T_{\text{out}}$ are averaged over mentions, then
macro-averaged over the nine ZELDA datasets or four ZESHEL domains. $T_{\text{out}}$ is the
mean over the ten samples.

\section{Pass@$k$ and Maj@$k$}
\label{sec:passk}

For test-time scaling methods, we sample $n$ completions per mention (we use
$n=10$), each producing a parsed answer index. From these we estimate two
metrics, Pass@$k$ and Maj@$k$, over subsets of the $n$ samples. Both are
computed only over mentions whose gold entity lies in the top-$k$ shortlist
$C_m^{(k)}$.

\paragraph{Pass@$k$.}
Given $n$ completions of which $c$ are correct, Pass@$k$ is the probability
that at least one of $k$ uniformly drawn completions (without replacement) is
correct. We use the unbiased estimator of~\citet{chenEvaluatingLargeLanguage2021}:
\begin{equation}
\mathrm{Pass}@k = 1 - \frac{\binom{n-c}{k}}{\binom{n}{k}}.
\end{equation}

\paragraph{Maj@$k$.}
To quantify how much of this coverage plain self-consistency
decoding~\cite{wangSelfConsistencyImprovesChain2023} recovers, we pair Pass@$k$
with Maj@k which we define as the expected majority-vote accuracy over uniformly sampled subsets. 
Let $j^\ast$ be the index of the gold entity in $C_m^{(k)}$, and let
$\hat\jmath^{(l)}\in\{1,\dots,k\}\cup\{\varnothing\}$ denote the parsed answer of
the $l$-th completion ($\varnothing$ for unparseable outputs). For a size-$k$
subset $S$ of the $n$ completions, the vote count for index $j$ is
$v_j(S)=\big|\{\,l\in S : \hat\jmath^{(l)}=j\,\}\big|$, and the (possibly tied)
set of winning indices is
$M(S)=\{\,j : v_j(S)=\max_{j'} v_{j'}(S)\,\}$, with $M(S)=\varnothing$ when no
completion in $S$ produces a valid index. We define
\begin{equation}
\mathrm{Maj}@k = \mathbb{E}_{S\sim\binom{[n]}{k}}
\!\left[\frac{1}{|M(S)|}\,\mathbf{1}\!\left[\,j^\ast \in M(S)\,\right]\right],
\end{equation}
where the expectation is taken uniformly over all $\binom{n}{k}$ subsets of size
$k$ and computed exactly by enumeration. The factor $1/|M(S)|$ scores ties as the
expected accuracy under uniform random tie-breaking, making Maj@$k$ an unbiased
estimate of majority-vote accuracy.

\paragraph{Properties.}
The gap $\mathrm{Pass}@k - \mathrm{Maj}@k$ measures the headroom left by plain
self-consistency: mentions on which a correct answer is present among the $k$
samples but majority voting fails to surface it. In particular, Pass@$n$
(here Pass@$10$) equals the fraction of mentions with at least one correct
completion, i.e.\ the ceiling achievable by any aggregation over the $n$ samples.

\section{Reranker-induced Upper-bounds}
\label{sec:acck}

Reranking accuracy@10 is reported in Table~\ref{tab:reranker_acc10} for the base and the (Wikipedia) fine-tuned variants. 

\begin{table}[t]
\centering
\small
\setlength{\tabcolsep}{2.0pt}
\begin{subtable}{\linewidth}
\centering
\begin{tabular}{lcccccc}
\toprule
 & \multicolumn{3}{c}{\textbf{Qwen3-Rerank-0.6B}} & \multicolumn{3}{c}{\textbf{Qwen3-Rerank-4B}} \\
\cmidrule(lr){2-4} \cmidrule(lr){5-7}
\textbf{Domain} & Base & FT & $\Delta$ & Base & FT & $\Delta$ \\
\midrule
AIDA-B             & \done{94.31} & \done{94.98} & +0.67 & \done{95.50} & \done{95.81} & +0.31 \\
Tweeki             & \done{90.58} & \done{93.26} & +2.68 & \done{93.26} & \done{94.19} & +0.93 \\
R.-Posts       & \done{95.73} & \done{98.29} & +2.56 & \done{98.29} & \done{98.72} & +0.43 \\
R.-Comments    & \done{94.51} & \done{97.65} & +3.14 & \done{97.49} & \done{97.81} & +0.32 \\
W.-CWEB          & \done{87.06} & \done{88.69} & +1.63 & \done{88.02} & \done{90.11} & +2.09 \\
W.-Wiki          & \done{96.58} & \done{98.03} & +1.45 & \done{97.94} & \done{98.67} & +0.73 \\
S.-Tail   & \done{97.44} & \done{97.44} & +0.00 & \done{97.44} & \done{97.44} & +0.00 \\
S.-Shadow & \done{57.19} & \done{57.41} & +0.22 & \done{57.52} & \done{57.74} & +0.22 \\
S.-Top    & \done{73.34} & \done{73.56} & +0.22 & \done{73.56} & \done{73.67} & +0.11 \\
\midrule
Macro avg.         & \done{87.42} & \done{88.81} & +1.40 & \done{88.78} & \done{89.35} & +0.57 \\
\bottomrule
\end{tabular}
\caption{ZELDA.}
\label{tab:reranker_acc10_zelda}
\end{subtable}

\vspace{\baselineskip}

\begin{subtable}{\linewidth}
\centering
\begin{tabular}{lcccccc}
\toprule
 & \multicolumn{3}{c}{\textbf{Qwen3-Reranker-0.6B}} & \multicolumn{3}{c}{\textbf{Qwen3-Reranker-4B}} \\
\cmidrule(lr){2-4} \cmidrule(lr){5-7}
\textbf{Domain} & Base & FT & $\Delta$ & Base & FT & $\Delta$ \\
\midrule
F. Realms & \done{77.00} & \done{97.60} & +20.60 & \done{97.80} & \done{98.70} & +0.90 \\
Lego             & \done{71.39} & \done{93.63} & +22.24 & \done{95.48} & \done{98.25} & +2.77 \\
Star Trek        & \done{54.74} & \done{93.23} & +38.49 & \done{94.02} & \done{97.52} & +3.50 \\
YuGiOh           & \done{45.94} & \done{94.53} & +48.59 & \done{92.28} & \done{98.53} & +6.25 \\
\midrule
Macro avg.       & \done{62.27} & \done{94.75} & +32.48 & \done{94.90} & \done{98.25} & +3.35 \\
\bottomrule
\end{tabular}
\caption{ZESHEL.}
\label{tab:reranker_acc10_zeshel}
\end{subtable}

\vspace{\baselineskip}

\begin{subtable}{\linewidth}
\centering
\begin{tabular}{lcccccc}
\toprule
 & \multicolumn{3}{c}{\textbf{Qwen3-Reranker-0.6B}} & \multicolumn{3}{c}{\textbf{Qwen3-Reranker-4B}} \\
\cmidrule(lr){2-4} \cmidrule(lr){5-7}
\textbf{Domain} & Base & FT & $\Delta$ & Base & FT & $\Delta$ \\
\midrule
Biomedical & \done{43.57} & \done{61.99} & +18.42 & \done{77.77} & \done{83.65} & +5.88 \\
\bottomrule
\end{tabular}
\caption{GLADIS.}
\label{tab:reranker_acc10_gladis}
\end{subtable}
\caption{Acc@10 of the base Qwen3-Reranker and our finetuned rerankers (\textsc{FT}) on ZELDA~\subref{tab:reranker_acc10_zelda} and GLADIS biomedical~\subref{tab:reranker_acc10_gladis},
         and normalized Acc@10 on ZESHEL~\subref{tab:reranker_acc10_zeshel}, defining the recall ceiling for all downstream
         selection-based methods. $\Delta$ denotes the improvement from finetuning (on Wikipedia).}
\label{tab:reranker_acc10}
\vspace{-\baselineskip}
\end{table}

\section{Additional results}
\label{sec:additional_results}

\paragraph{Complementary results.} Additional baseline and cross-family replication results in the classical, true zero-shot and few-shot settings are reported, respectively, in Table~\ref{tab:zelda_crossfamily}, Table~\ref{tab:zeshel_crossfamily} and Table~\ref{tab:few-shot_additional_results}.

\begin{table*}
  \centering

\begin{subtable}{\textwidth}
    \centering
    \small
    \setlength{\tabcolsep}{2pt}
    \begin{tabular}{lcccccccccc}
      \toprule
      Model & AIDA & Tweeki & \multicolumn{2}{c}{Reddit} & \multicolumn{2}{c}{WNED} & \multicolumn{3}{c}{Slinks} & Avg \\
      \cmidrule(lr){4-5}\cmidrule(lr){6-7}\cmidrule(lr){8-10}
      & & & posts & comments & CWEB & Wiki & Tail & Shadow & Top & \\
      \midrule
      \multicolumn{11}{c}{\textbf{Baselines}} \\
      \midrule
      Upper bound & 91.1 & 94.0 & 98.4 & 98.3 & 92.4 & 98.8 & 98.8 & 56.7 & 73.1 & 89.1 \\
      MFS baseline & 63.5 & 72.3 & 83.4 & 81.0 & 61.2 & 65.1 & 99.4 & 14.9 & 41.3 & 64.7 \\
      $\text{FEVRY}_\text{CL}$ & 79.5 & 76.9 & 89.0 & 86.5 & 70.3 & 84.5 & 87.6 & 31.9 & 47.7 & 72.7 \\
      $\text{LUKE}_\text{FT}$ & 81.2 & 77.9 & 81.5 & 78.5 & 70.3 & 76.5 & 98.0 & 22.5 & 51.8 & 71.0 \\
      $\text{GENRE}_\text{CL}$ & 78.6 & 80.1 & 92.8 & 91.5 & 73.6 & 88.4 & 99.6 & 37.3 & 52.8 & 77.2 \\
      \multicolumn{11}{l}{LELA w/} \\
      \rowcolor{lightgray} \texttt{Qwen3-30B-A3B} & \ci{85.2}{1.0} & \ci{78.7}{2.7} & \ci{82.8}{2.8} & \ci{84.8}{2.8} & \ci{70.0}{0.9} & \ci{88.1}{0.8} & \ci{97.2}{1.1} & \ci{50.4}{3.3} & \ci{68.1}{3.0} & \ci{78.4}{0.8} \\
      \texttt{gemma-4-31B-it} & \ci{87.0}{1.0} & \ci{81.2}{2.6} & \ci{88.5}{2.4} & \ci{87.0}{2.6} & \ci{71.5}{0.8} & \ci{89.6}{0.7} & \ci{97.2}{1.1} & \ci{50.7}{3.3} & \ci{67.5}{3.1} & \ci{80.0}{0.7} \\
      \texttt{gemma-4-E2B-it} & \ci{82.2}{1.1} & \ci{76.3}{2.8} & \ci{84.4}{2.7} & \ci{82.8}{2.9} & \ci{68.6}{0.9} & \ci{83.5}{0.9} & \ci{97.1}{1.1} & \ci{45.7}{3.3} & \ci{64.2}{3.1} & \ci{76.1}{0.8} \\
      \texttt{gemma-4-E4B-it} & \ci{83.3}{1.1} & \ci{77.1}{2.8} & \ci{81.5}{2.9} & \ci{78.7}{3.2} & \ci{67.8}{0.9} & \ci{86.2}{0.8} & \ci{97.1}{1.1} & \ci{48.6}{3.3} & \ci{64.3}{3.1} & \ci{76.1}{0.8} \\
      \midrule
      \multicolumn{11}{c}{\textbf{Selection (Test Time Scaling)}} \\
      \midrule
      \multicolumn{11}{c}{\textbf{RLVR}} \\
      \texttt{gemma-4-E2B-it} & \done{\ci{86.8}{1.0}} & \done{\ci{80.7}{2.6}} & \done{\ci{92.2}{2.0}} & \done{\ci{92.3}{2.1}} & \done{\ci{73.7}{0.8}} & \done{\ci{86.7}{0.8}} & \done{\ci{97.2}{1.1}} & \done{\ci{42.5}{3.2}} & \done{\ci{60.3}{3.2}} & \done{\ci{79.2}{0.7}} \\
      \texttt{gemma-4-E4B-it} & \done{\ci{87.9}{1.0}} & \done{\ci{82.1}{2.6}} & \done{\ci{91.6}{2.1}} & \done{\ci{91.5}{2.2}} & \done{\ci{74.0}{0.8}} & \done{\ci{89.2}{0.7}} & \done{\ci{97.2}{1.1}} & \done{\ci{45.6}{3.3}} & \done{\ci{64.1}{3.1}} & \done{\ci{80.4}{0.7}} \\
      \midrule
      \multicolumn{11}{c}{\textbf{On-Policy Distillation}} \\
      \texttt{Qwen3-0.6B} & \done{\ci{82.7}{1.1}} & \done{\ci{80.5}{2.7}} & \done{\ci{91.9}{2.0}} & \done{\ci{90.9}{2.2}} & \done{\ci{73.0}{0.8}} & \done{\ci{86.5}{0.8}} & \done{\ci{97.3}{1.1}} & \done{\ci{43.5}{3.2}} & \done{\ci{59.0}{3.2}} & \done{\ci{78.4}{0.7}} \\
      \texttt{Qwen3-1.7B} & \done{\ci{84.2}{1.1}} & \done{\ci{80.1}{2.7}} & \done{\ci{88.3}{2.4}} & \done{\ci{88.4}{2.5}} & \done{\ci{72.0}{0.8}} & \done{\ci{87.1}{0.8}} & \done{\ci{97.2}{1.1}} & \done{\ci{44.8}{3.3}} & \done{\ci{61.6}{3.2}} & \done{\ci{78.2}{0.7}} \\
      \done{\texttt{Qwen3-4B}} & \done{\ci{84.7}{1.1}} & \done{\ci{79.9}{2.7}} & \done{\ci{81.8}{2.9}} & \done{\ci{84.2}{2.8}} & \done{\ci{70.5}{0.9}} & \done{\ci{87.8}{0.8}} & \done{\ci{97.3}{1.1}} & \done{\ci{48.9}{3.3}} & \done{\ci{65.6}{3.1}} & \done{\ci{77.9}{1.2}} \\
      \midrule
      \multicolumn{11}{c}{\textbf{Off-Policy Distillation}} \\
      \texttt{Qwen3-0.6B} & \done{\ci{84.2}{1.1}} & \done{\ci{82.1}{2.6}} & \done{\ci{92.9}{1.9}} & \done{\ci{92.0}{2.1}} & \done{\ci{73.0}{0.8}} & \done{\ci{86.9}{0.8}} & \done{\ci{97.3}{1.1}} & \done{\ci{43.6}{3.2}} & \done{\ci{59.5}{3.2}} & \done{\ci{79.1}{0.7}} \\
      \texttt{Qwen3-1.7B} & \done{\ci{86.0}{1.0}} & \done{\ci{80.9}{2.6}} & \done{\ci{89.8}{2.2}} & \done{\ci{89.8}{2.4}} & \done{\ci{72.7}{0.8}} & \done{\ci{87.4}{0.8}} & \done{\ci{97.3}{1.1}} & \done{\ci{45.4}{3.3}} & \done{\ci{61.5}{3.2}} & \done{\ci{79.0}{0.7}} \\ 
      \done{\texttt{Qwen3-4B}} & \done{\ci{86.3}{1.0}} & \done{\ci{80.0}{2.7}} & \done{\ci{85.1}{2.6}} & \done{\ci{86.5}{2.7}} & \done{\ci{71.7}{0.8}} & \done{\ci{88.4}{0.8}} & \done{\ci{97.3}{1.1}} & \done{\ci{49.0}{3.3}} & \done{\ci{66.3}{3.1}} & \done{\ci{79.0}{0.7}} \\
      \midrule
      \multicolumn{11}{c}{\textbf{Off-Policy Distillation + RLVR}} \\
      \texttt{Qwen3-0.6B} & \done{\ci{85.1}{1.0}} & \done{\ci{81.6}{2.6}} & \done{\ci{93.6}{1.8}} & \done{\ci{93.7}{1.9}} & \done{\ci{73.9}{0.8}} & \done{\ci{88.6}{0.8}} & \done{\ci{97.2}{1.1}} & \done{\ci{42.2}{3.2}} & \done{\ci{58.0}{3.2}} & \done{\ci{79.3}{0.7}} \\
      \texttt{gemma-4-E2B-it} & \done{\ci{86.2}{1.0}} & \done{\ci{82.8}{2.5}} & \done{\ci{91.5}{2.1}} & \done{\ci{93.0}{2.0}} & \done{\ci{73.5}{0.8}} & \done{\ci{87.7}{0.8}} & \done{\ci{97.2}{1.1}} & \done{\ci{42.7}{3.2}} & \done{\ci{61.2}{3.2}} & \done{\ci{79.5}{0.7}} \\
    \texttt{gemma-4-E4B-it} & \done{\ci{87.9}{1.0}} & \done{\ci{82.3}{2.6}} & \done{\ci{93.3}{1.9}} & \done{\ci{91.9}{2.1}} & \done{\ci{74.6}{0.8}} & \done{\ci{89.1}{0.8}} & \done{\ci{97.2}{1.1}} & \done{\ci{45.2}{3.3}} & \done{\ci{63.6}{3.1}} & \done{\ci{80.6}{0.7}} \\
      \bottomrule
    \end{tabular}
    \caption{Additional results on ZELDA.}
    \label{tab:zelda_crossfamily}
  \end{subtable}

  \vspace{0.5\baselineskip}

  \begin{subtable}{\textwidth}
    \centering
    \small
    \begin{tabular}{lccccc}
      \toprule
      \textbf{Method} & \textbf{F. Realms} & \textbf{Lego} & \textbf{Star Trek} & \textbf{YuGiOh} & \textbf{Macro Acc.} \\
      \midrule
      \multicolumn{6}{c}{\textbf{Baselines}} \\
      \midrule
      ExtEnD~\cite{barbaExtEnDExtractiveEntity2022} & 79.62 & 65.20 & 73.21 & 60.01 & 69.51 \\
      GENRE~\cite{decaoAutoregressiveEntityRetrieval2021} & 55.20 & 42.71 & 55.76 & 34.68 & 47.09 \\
      OneNet~\cite{liuOneNetFineTuningFree2024a} & 55.80 & 53.80 & 53.90 & 40.80 & 51.08 \\
      LELA~\cite{haffoudhiLELALLMBasedEntity2026a} \\
      \rowcolor{lightgray} w/ \texttt{Qwen3-30B-A3B-Thinking-2507} & \ci{87.90}{2.02} & \ci{81.21}{2.46} & \ci{76.95}{1.57} & \ci{77.45}{1.81} & \ci{80.88}{0.99} \\
      w/ \texttt{gemma-4-31B-it} & \ci{92.60}{1.63} & \ci{87.58}{2.08} & \ci{84.96}{1.33} & \ci{86.70}{1.47} & \ci{87.96}{0.83} \\
      w/ \texttt{gemma-4-E2B-it} & \ci{63.06}{3.00} & \ci{50.98}{3.15} & \ci{53.11}{1.85} & \ci{41.26}{2.13} & \ci{52.10}{1.30} \\
      w/ \texttt{gemma-4-E4B-it} & \ci{80.28}{2.47} & \ci{73.59}{2.78} & \ci{63.41}{1.79} & \ci{55.82}{2.15} & \ci{68.28}{1.16} \\
      \midrule
      \multicolumn{6}{c}{\textbf{Selection (Test Time Scaling)}} \\
      \midrule
      \multicolumn{6}{c}{\textbf{RLVR}} \\
      \texttt{gemma-4-E2B-it} & \done{\ci{81.30}{2.42}} & \done{\ci{68.99}{2.91}} & \done{\ci{67.72}{1.74}} & \done{\ci{60.59}{2.12}} & \done{\ci{69.65}{1.17}} \\
      \texttt{gemma-4-E4B-it} & \done{\ci{83.90}{2.28}} & \done{\ci{79.06}{2.56}} & \done{\ci{76.73}{1.57}} & \done{\ci{65.37}{2.06}} & \done{\ci{76.27}{1.07}} \\
      \midrule
      \multicolumn{6}{c}{\textbf{On-Policy Distillation}} \\
      \texttt{Qwen3-0.6B} & \done{\ci{74.50}{2.71}} & \done{\ci{56.98}{3.11}} & \done{\ci{62.41}{1.80}} & \done{\ci{59.91}{2.12}} & \done{\ci{63.45}{1.24}} \\
      \texttt{Qwen3-1.7B} & \done{\ci{83.30}{2.32}} & \done{\ci{71.66}{2.84}} & \done{\ci{73.11}{1.65}} & \done{\ci{71.85}{1.95}} & \done{\ci{74.98}{1.12}} \\
      \done{\texttt{Qwen3-4B}} & \done{\ci{86.00}{2.15}} & \done{\ci{76.49}{2.67}} & \done{\ci{74.98}{1.31}} & \done{\ci{77.20}{1.82}} & \done{\ci{78.67}{1.02}} \\
      \midrule
      \multicolumn{6}{c}{\textbf{Off-Policy Distillation}} \\
      \texttt{Qwen3-0.6B} & \done{\ci{82.60}{2.35}} & \done{\ci{68.79}{2.92}} & \done{\ci{72.68}{1.66}} & \done{\ci{68.29}{2.01}} & \done{\ci{73.09}{1.14}} \\
      \texttt{Qwen3-1.7B} & \done{\ci{83.70}{2.29}} & \done{\ci{72.59}{2.81}} & \done{\ci{74.69}{1.62}} & \done{\ci{72.77}{1.93}} & \done{\ci{75.94}{1.10}} \\
      \done{\texttt{Qwen3-4B}} & \done{\ci{86.80}{2.10}} & \done{\ci{79.77}{2.53}} & \done{\ci{78.49}{1.24}} & \done{\ci{78.47}{1.78}} & \done{\ci{80.88}{0.99}} \\
      \midrule
      \multicolumn{6}{c}{\textbf{Off-Policy Distillation + RLVR}} \\
      \texttt{Qwen3-0.6B} & \done{\ci{79.00}{2.53}} & \done{\ci{64.07}{3.02}} & \done{\ci{70.20}{1.70}} & \done{\ci{63.96}{2.08}} & \done{\ci{69.31}{1.19}} \\
      \texttt{gemma-4-E2B-it} & \done{\ci{81.60}{2.41}} & \done{\ci{69.30}{2.90}} & \done{\ci{71.78}{1.67}} & \done{\ci{69.07}{2.00}} & \done{\ci{72.94}{1.15}} \\
      \texttt{gemma-4-E4B-it} & \done{\ci{84.60}{2.24}} & \done{\ci{77.72}{2.62}} & \done{\ci{75.01}{1.61}} & \done{\ci{71.54}{1.99}} & \done{\ci{77.22}{1.07}} \\
      \bottomrule
    \end{tabular}
    \caption{Additional results on the ZESHEL test set (true zero-shot).}
    \label{tab:zeshel_crossfamily}
  \end{subtable}

    \caption{Additional results in the classical (\subref{tab:zelda_crossfamily})
    and true zero-shot (\subref{tab:zeshel_crossfamily}) settings.}  \label{tab:crossfamily}
\end{table*}

\begin{table*}
  \centering
  \small
  \begin{tabular}{lccccc}
    \toprule
    \textbf{Method} & \textbf{F. Realms} & \textbf{Lego} & \textbf{Star Trek} & \textbf{YuGiOh} & \textbf{Macro Acc.} \\
    \midrule
    \multicolumn{6}{c}{\textbf{Baselines}} \\
    \midrule
    LELA~\cite{haffoudhiLELALLMBasedEntity2026a} \\
    \rowcolor{lightgray} w/ \texttt{gemma-4-31B-it} & \ci{92.62}{1.69} & \ci{87.22}{2.20} & \ci{84.95}{1.34} & \ci{86.64}{1.50} & \ci{87.86}{0.86} \\
    \midrule
    \multicolumn{6}{c}{\textbf{RLVR} \textrightarrow{} \textbf{RLVR}} \\
    \texttt{Qwen3-4B} & \done{\ci{86.86}{2.19}} & \done{\ci{79.48}{2.66}} & \done{\ci{79.41}{1.52}} & \done{\ci{81.42}{1.71}} & \done{\ci{81.79}{1.03}} \\
    \midrule
    \multicolumn{6}{c}{\textbf{Off-Policy Distillation} \textrightarrow{} \textbf{On-Policy Distillation} (\texttt{gemma-4-31B-it} teacher in-domain)} \\
    \texttt{gemma-4-E4B-it} & \done{\underline{\ci{88.38}{2.07}}} & \done{\ci{80.72}{2.59}} & \done{\ci{79.11}{1.53}} & \done{\ci{82.52}{1.67}} & \done{\ci{82.68}{1.00}} \\
    \multicolumn{6}{c}{\textbf{Off-Policy Distillation} \textrightarrow{} \textbf{Off-Policy Distillation} (\texttt{gemma-4-31B-it} teacher in-domain)} \\
    \texttt{gemma-4-E4B-it} & \done{\underline{\ci{87.95}{2.11}}} & \done{\underline{\ci{82.06}{2.52}}} & \done{\ci{79.55}{1.52}} & \done{\ci{82.47}{1.67}} & \done{\ci{83.01}{1.00}} \\
    \bottomrule
  \end{tabular}
  \caption{Results on ZESHEL, in the few-shot setting, for additional paradigm combinations.
  }
  \label{tab:few-shot_additional_results}
  \vspace{-\baselineskip}
\end{table*}

\section{Category-specific Results}
\label{app:categories}

In ZESHEL, mentions are split into four categories, based on token
overlap with the entity name:

\begin{description}
\item [High Overlap (HO):] the entity name is identical to the mention text
    \item[Multiple Categories (MC):] the entity name is the mention text followed by a
        disambiguation phrase (e.g. mention string: ``Batman'', title: ``Batman
        (Lego)'')
    \item[Ambiguous Substrings (AS):] the mention text is a substring of the entity name
    \item[Low Overlap (LO):] all other cases
\end{description}

\noindent These categories respectively make up approximately 5\%, 28\%, 8\% and 59\% of the mentions
in the dataset. 
Results are reported in Table~\ref{tab:zeshel_category_results}.

\noindent At 4B, off-policy distillation improves over the reranker by $8.7$ points
on \textsc{mc} against $4.0$ on \textsc{lo} and at most $1.2$ elsewhere.
\textsc{mc} is precisely the category in which every candidate shares the
mention's surface form and differs only by a disambiguation phrase, so
lexical overlap carries no discriminative signal, making it the lexical
counterpart of a low reranker margin (Appendix~\ref{app:complementarity}).
It is also the one category on which scaling the reranker from 0.6B to 4B
buys little, and the one on which the distilled 4B student remains furthest from its teacher,
though it exceeds it on \textsc{ho}. Reasoning-path gain is thus localized along a lexical axis as well as a logit one.

\begin{table*}
  \centering
  \small
  \setlength{\tabcolsep}{4pt}
  \begin{tabular}{lcccc}
    \toprule
    \textbf{Method} & \textbf{LO} & \textbf{MC} & \textbf{AS} & \textbf{HO} \\
    \midrule
    \multicolumn{5}{c}{\textbf{Baselines}} \\
    \midrule
    LELA \\
    \rowcolor{lightgray} w/ \texttt{gemma-4-26B-A4B-it} & \ci{78.83}{1.52} & \ci{88.78}{1.41} & \ci{89.61}{3.06} & \ci{95.44}{1.93} \\
    \midrule
    \multicolumn{5}{c}{\textbf{Pointwise Reranking}} \\
    \midrule
    \texttt{Qwen3-Reranker-0.6B} & \done{\ci{56.14}{1.84}} & \done{\ci{69.67}{2.14}} & \done{\ci{75.59}{4.53}} & \done{\ci{90.91}{2.59}} \\
    \texttt{Qwen3-Reranker-4B}   & \done{\ci{73.91}{1.64}} & \done{\ci{75.54}{1.97}} & \done{\ci{86.95}{3.50}} & \done{\ci{97.52}{1.40}} \\
    \midrule
    \multicolumn{5}{c}{\textbf{Direct Selection}} \\
    \midrule
    \texttt{Qwen3-0.6B} & \done{\ci{49.74}{1.73}} & \done{\ci{63.26}{1.95}} & \done{\ci{65.16}{4.58}} & \done{\ci{83.03}{2.57}} \\
    \texttt{Qwen3-1.7B} & \done{\ci{62.78}{1.67}} & \done{\ci{66.64}{1.91}} & \done{\ci{74.22}{4.21}} & \done{\ci{90.30}{2.02}} \\
    \texttt{Qwen3-4B}   & \done{\ci{70.62}{1.57}} & \done{\ci{76.21}{1.73}} & \done{\ci{81.86}{3.70}} & \done{\ci{93.09}{1.73}} \\
    \midrule
    \multicolumn{5}{c}{\textbf{Selection (Test Time Scaling)}} \\
    \midrule
    \multicolumn{5}{c}{\textbf{RLVR}} \\
    \texttt{Qwen3-0.6B} & \done{\ci{37.56}{1.67}} & \done{\ci{51.39}{2.03}} & \done{\ci{61.58}{4.68}} & \done{\ci{73.33}{3.02}} \\
    \texttt{Qwen3-1.7B} & \done{\ci{64.24}{1.65}} & \done{\ci{69.67}{1.86}} & \done{\ci{82.10}{3.69}} & \done{\ci{95.76}{1.38}} \\
    \texttt{Qwen3-4B}   & \done{\ci{69.76}{1.59}} & \done{\ci{79.50}{1.64}} & \done{\ci{84.96}{3.44}} & \done{\ci{97.58}{1.05}} \\
    \multicolumn{5}{c}{\textbf{On-Policy Distillation}} \\
    \texttt{gemma-4-E2B-it} & \done{\ci{61.11}{1.68}} & \done{\ci{67.88}{1.89}} & \done{\ci{74.70}{4.18}} & \done{\ci{88.24}{2.20}} \\
    \texttt{gemma-4-E4B-it} & \done{\ci{71.40}{1.56}} & \done{\ci{77.45}{1.69}} & \done{\ci{82.34}{3.67}} & \done{\ci{94.42}{1.57}} \\
    \multicolumn{5}{c}{\textbf{Off-Policy Distillation}} \\
    \texttt{Qwen3-0.6B} & \done{\cibf{66.38}{1.63}} & \done{\cibf{75.57}{1.74}} & \done{\cibf{81.62}{3.72}} & \done{\cibf{95.64}{1.40}} \\
    \texttt{gemma-4-E2B-it} & \done{\cibf{70.19}{1.58}} & \done{\cibf{80.05}{1.62}} & \done{\cibf{82.82}{3.63}} & \done{\cibf{96.97}{1.17}} \\
    \texttt{gemma-4-E4B-it} & \done{\cibf{77.94}{1.43}} & \done{\cibf{84.28}{1.48}} & \done{\cibf{87.11}{3.22}} & \done{\cibf{98.67}{0.78}} \\
    \bottomrule
  \end{tabular}
  \caption{Per-category normalized accuracy on the ZESHEL test set. \textbf{Bold}: best
           per size group per column.}
  \label{tab:zeshel_category_results}
  \vspace{-\baselineskip}
\end{table*}

\section{True Zero-Shot Results on GLADIS}
\label{app:gladis}

\paragraph{Benchmark.} To broaden both the domain and the task, we evaluate on 3,149 biomedical acronym-disambiguation examples from GLADIS~\cite{chenGLADISGeneralLarge2023}. Mentions are acronyms in sentences, and entities are their long forms, drawn from GLADIS's acronym dictionary, which lists 278 long forms per acronym on average.

\paragraph{Configuration.} Following \citet{haffoudhiLELALLMBasedEntity2026a}, all long forms of an acronym are its candidates, with no description, and we report accuracy.
As for ZESHEL, we reuse the checkpoints from Section~\ref{sec:classical} and evaluate them directly, matching the true zero-shot setting.
The shortlist is less complete than on the other benchmarks: the Acc@10 of the fine-tuned reranker is 83.7\% at 4B and 62.0\% at 0.6B (Table~\ref{tab:reranker_acc10_gladis}, Appendix~\ref{sec:acck}), which caps all listwise paradigms.

\paragraph{Results.} Results are reported in Table~\ref{tab:gladis_results}.
The ordering observed on ZESHEL holds: off-policy distillation is best in every size group, and direct selection beats pointwise reranking at every scale.
At 4B, reasoning's margin over the reranker grows from about 6 points on ZESHEL to over 20.
Here, training helps at smaller scales (from 10.7\% to 26.7\% at 0.6B and from
20.1\% to 36.6\% in the 1.7B/2B group), but LELA using \texttt{Qwen3-4B} matches
the best trained 4B model, and the teacher remains ahead: lacking entity
descriptions and posing a different task, GLADIS calls for highly specific
in-domain knowledge, which ED training on Wikipedia does not provide, rather
than ED-specific reasoning capabilities.

\begin{table}[t]
  \centering
  \small
  \begin{tabular}{lc}
    \toprule
    \textbf{Method} & \textbf{Acc.} \\
    \midrule
    \multicolumn{2}{c}{\textbf{Baselines}} \\
    \midrule
    BLINK~\cite{wuScalableZeroshotEntity2020} & 14.34 \\
    LELA~\cite{haffoudhiLELALLMBasedEntity2026a} \\
    \rowcolor{lightgray} w/ \texttt{gemma-4-26B-A4B-it} & \ci{65.74}{1.80} \\
    w/ \texttt{Qwen3-0.6B} & \ci{10.67}{1.08} \\
    w/ \texttt{Qwen3-1.7B} & \ci{20.10}{1.40} \\
    w/ \texttt{Qwen3-4B}   & \ci{53.51}{1.74} \\
    \midrule
    \multicolumn{2}{c}{\textbf{Pointwise Reranking}} \\
    \midrule
    \texttt{Qwen3-Reranker-0.6B} & \ci{13.62}{1.20} \\
    \texttt{Qwen3-Reranker-4B}   & \ci{30.61}{1.61} \\
    \midrule
    \multicolumn{2}{c}{\textbf{Direct Selection}} \\
    \midrule
    \texttt{Qwen3-0.6B} & \ci{17.12}{1.32} \\
    \texttt{Qwen3-1.7B} & \ci{24.64}{1.51} \\
    \texttt{Qwen3-4B}   & \ci{41.47}{1.72} \\
    \midrule
    \multicolumn{2}{c}{\textbf{Selection (Test Time Scaling)}} \\
    \midrule
    \multicolumn{2}{c}{\textbf{RLVR}} \\
    \texttt{Qwen3-0.6B} & \ci{17.62}{1.33} \\
    \texttt{Qwen3-1.7B} & \ci{33.22}{1.65} \\
    \texttt{Qwen3-4B}   & \ci{48.17}{1.75} \\
    \multicolumn{2}{c}{\textbf{On-Policy Distillation}} \\
    \texttt{gemma-4-E2B-it} & \ci{33.41}{1.65} \\
    \texttt{gemma-4-E4B-it} & \ci{48.65}{1.75} \\
    \multicolumn{2}{c}{\textbf{Off-Policy Distillation}} \\
    \texttt{Qwen3-0.6B}     & \cibf{26.68}{1.55} \\
    \texttt{gemma-4-E2B-it} & \cibf{36.58}{1.68} \\
    \texttt{gemma-4-E4B-it} & \cibf{53.35}{1.74} \\
    \bottomrule
  \end{tabular}
  \caption{Accuracy on the biomedical domain of GLADIS in the true zero-shot setting. The best configuration is significantly better than every other in its group (baselines excluded).}
  \label{tab:gladis_results}
  \vspace{-\baselineskip}
\end{table}

\section{Complementarity and Routing}
\label{app:complementarity}

\paragraph{Setup.}
All of the following analyses compare \texttt{Qwen3-Reranker-4B} against the best
reasoning variant at 4B in each setting, restricted to mentions
whose gold entity is in the candidate shortlist.
Margin-conditioned analyses additionally require at least two
candidates, excluding $22.7\%$ of ZELDA mentions.
These excluded mentions contain no disagreements, so the
margin-conditioned results use a different denominator from
Table~\ref{tab:complementarity-main}.

\paragraph{Metrics.}
Let $g$ denote the fraction of mentions on which the reranker is
wrong and reasoning is correct, and $l$ the fraction on which
the reranker is correct and reasoning is wrong.
Their sum measures disagreement in correctness; it excludes
mentions where both systems are wrong, even if their predictions
differ. An oracle choosing the correct system whenever possible
achieves the reranker's accuracy plus $g$.
Table~\ref{tab:complementarity} supplements
Table~\ref{tab:complementarity-main} with standalone and oracle
accuracies, shared errors, and margin-based discrimination metrics.
AUC$_{\text{corr}}$ measures how well $z_1-z_2$ predicts reranker
correctness; AUC$_{\text{route}}$ measures how well it predicts
that the reranker is correct among mentions where exactly one
system is correct.

\paragraph{Margin standardization.}
We use the margin bins shown in
Figure~\ref{fig:margin_mediation}.
Let $w_b^{(s)}$ be the fraction of eligible mentions in setting
$s$ falling in bin $b$, and $d_b^{(s)}$ the disagreement rate
within that bin. The observed rate is
$d^{(s)}=\sum_b w_b^{(s)}d_b^{(s)}$.
Replacing each setting's bin weights with the classical weights
gives
\begin{equation}
\tilde{d}^{(s)}
= \sum_b w_b^{\mathrm{cls}} d_b^{(s)}.
\label{eq:standardisation}
\end{equation}
This compares settings under a common margin distribution while
preserving their within-bin disagreement rates.
We apply the same weights separately to the within-bin repair
and break rates, yielding $\tilde{g}^{(s)}$ and
$\tilde{l}^{(s)}$, and report their ratio
$\tilde{g}^{(s)}/\tilde{l}^{(s)}$.
This is the ratio of the standardized rates, not an average of
within-bin ratios.

The bin-level comparisons also reveal residual differences:
true zero-shot disagreement is lower than classical disagreement
in every bin above the lowest, whereas few-shot disagreement is
higher in four of the six bins.

\paragraph{Routing sweep.}
For each threshold, we use reasoning below the threshold and the
reranker's prediction otherwise.
We select a separate threshold for each evaluation split,
maximizing accuracy on that split itself.
This gives an optimistic bound for this threshold-routing rule,
rather than an estimate of a deployable router's performance.
The rule recovers $9$--$40\%$ of the accuracy gap between the
better standalone method and the oracle.
In one of the four true-zero-shot domains, the optimal threshold
routes every mention to reasoning, providing no inference saving
over always using reasoning.

\paragraph{Margin comparability.}
Interpreting Eq.~\ref{eq:standardisation} requires margin bins to
represent reasonably comparable levels of reranker uncertainty
across settings.
As a diagnostic, we compare reranker top-1 accuracy within each
bin with its classical value.
The mass-weighted average deviation is $2.5$ percentage points
in few-shot and $4.0$ in true zero-shot.
Low-margin bins are not systematically easier under shift,
providing evidence against a simple compression of the reranker's
logits as the explanation for the changed margin distribution.

\begin{table*}
\centering
\small
\begin{tabular}{lrrrrrr}
\toprule
Setting & Rerank & TTS & Oracle & Shared
        & AUC$_{\text{corr}}$ & AUC$_{\text{route}}$ \\
\midrule
Classical      & 89.49 & 89.31 & 92.35 & 52.9 & 0.853 & 0.605 \\
Few-shot       & 83.77 & 85.92 & 90.41 & 46.3 & 0.828 & 0.676 \\
True zero-shot & 77.64 & 83.69 & 86.88 & 52.6 & 0.778 & 0.627 \\
\bottomrule
\end{tabular}
\caption{Additional complementarity metrics at 4B, on mentions
whose gold entity is in the candidate shortlist.
Rerank, TTS, and Oracle are accuracies (\%).
Oracle selects the correct system whenever either is correct.
Shared is the percentage of errors made by both systems.
AUC$_{\text{corr}}$ measures margin-based prediction of reranker
correctness; AUC$_{\text{route}}$ measures prediction of which
system is correct when exactly one is correct.
Repair and break rates are reported in
Table~\ref{tab:complementarity-main}.}
\label{tab:complementarity}
\end{table*}

\section{Test-Time Scaling Behavior}
\label{app:tts}

\begin{figure}[t]
    \centering
    \includegraphics[width=.9\linewidth]{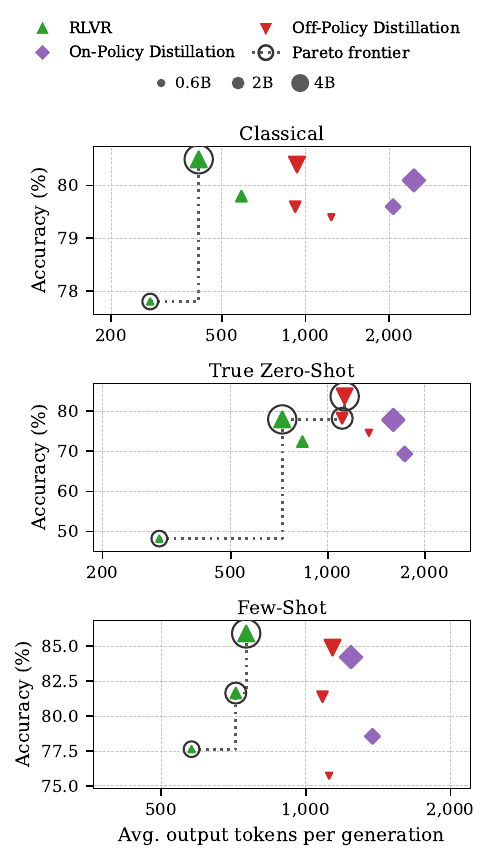}
    \caption{Accuracy vs.\ average output tokens per generation (log scale), per setting, for the
    TTS paradigms.}
    \label{fig:efficiency-tokens}
    \vspace{-\baselineskip}
\end{figure}

\begin{figure*}
    \centering

    \begin{subfigure}{0.49\linewidth}
        \includegraphics[width=\linewidth]{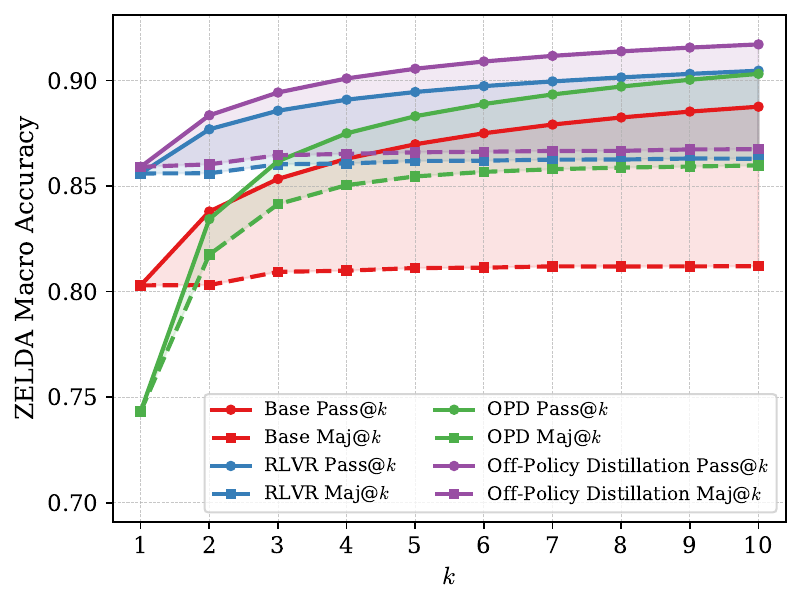}
        \caption{ZELDA (classical), 4B}
    \end{subfigure}
    \hfill
    \begin{subfigure}{0.49\linewidth}
        \includegraphics[width=\linewidth]{figures/zeshel_passmaj_gemma-4-e4b-it.pdf}
        \caption{ZESHEL (true zero-shot), 4B}
    \end{subfigure}
    
    \begin{subfigure}{0.49\linewidth}
        \includegraphics[width=\linewidth]{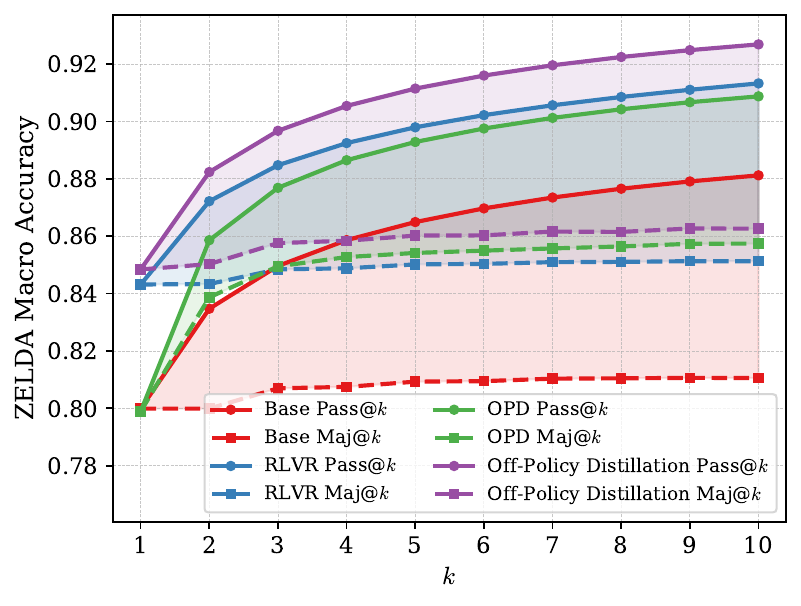}
        \caption{ZELDA (classical), 2B}
    \end{subfigure}
    \hfill
    \begin{subfigure}{0.49\linewidth}
        \includegraphics[width=\linewidth]{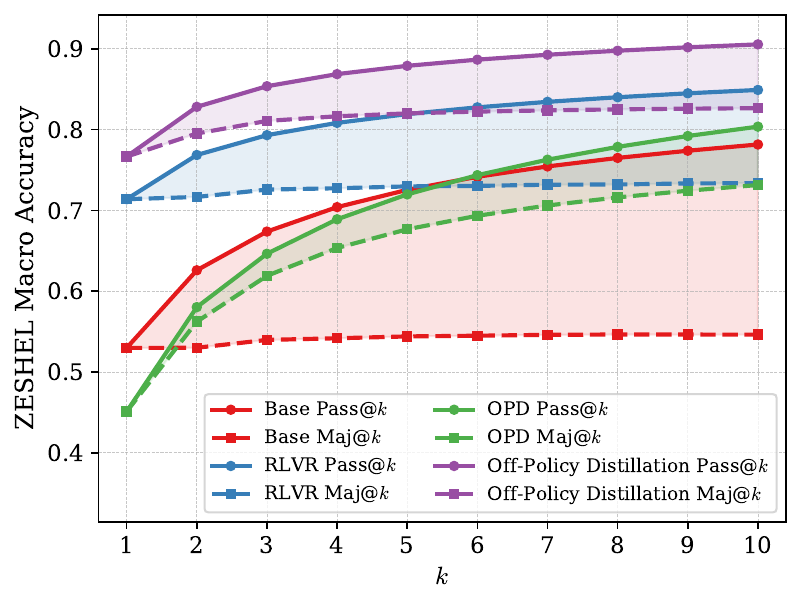}
        \caption{ZESHEL (true zero-shot), 2B}
    \end{subfigure}

    \begin{subfigure}{0.49\linewidth}
        \includegraphics[width=\linewidth]{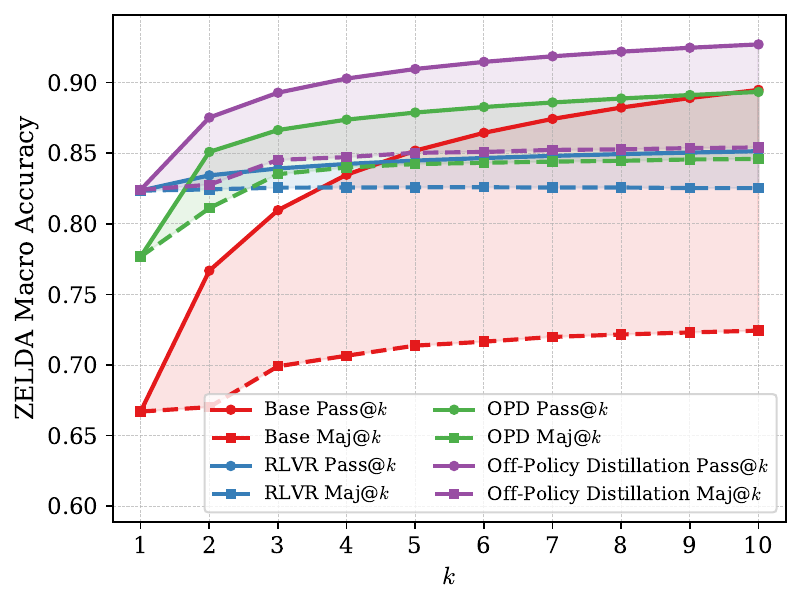}
        \caption{ZELDA (classical), 0.6B}
    \end{subfigure}
    \hfill
    \begin{subfigure}{0.49\linewidth}
        \includegraphics[width=\linewidth]{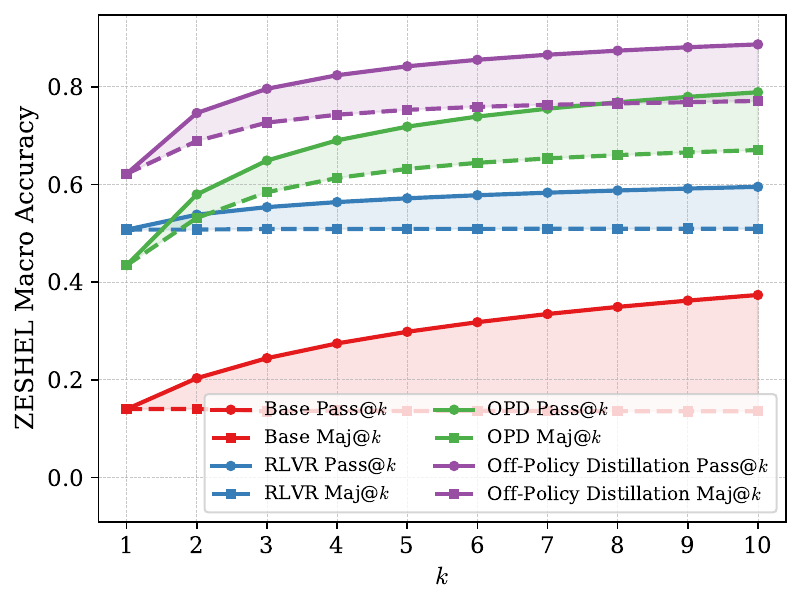}
        \caption{ZESHEL (true zero-shot), 0.6B}
    \end{subfigure}
    \caption{Pass@k and Maj@k over $k=1,\dots,10$.}
    \label{fig:passk}
    \vspace{-\baselineskip}
\end{figure*}

Figure~\ref{fig:passk} plots Pass@$k$ and Maj@$k$ against $k$ for each
training paradigm. The gap between them is the headroom left by plain
self-consistency, and its behavior separates the paradigms sharply.
Under RLVR, Maj@$k$ is nearly flat even when Pass@$k$ rises:
additional samples contain correct answers, but majority voting
rarely recovers them. Under distillation Pass@$k$ continues to rise with $k$ and
Maj@$k$ tracks it, so the additional samples are doing work. This is the
sampling-sharpening effect of RLVR~\citep{yueDoesReinforcementLearning2025a}, 
measured here on ED: RLVR
raises Pass@$1$ towards the base model's Pass@$10$ without extending it,
whereas distillation moves the ceiling itself.
Figure~\ref{fig:efficiency-tokens} plots accuracy against average output
tokens per generation. Distilled models emit more tokens than the
RLVR models at equal scale.

\begin{figure*}
    \centering
    \includegraphics[width=\linewidth]{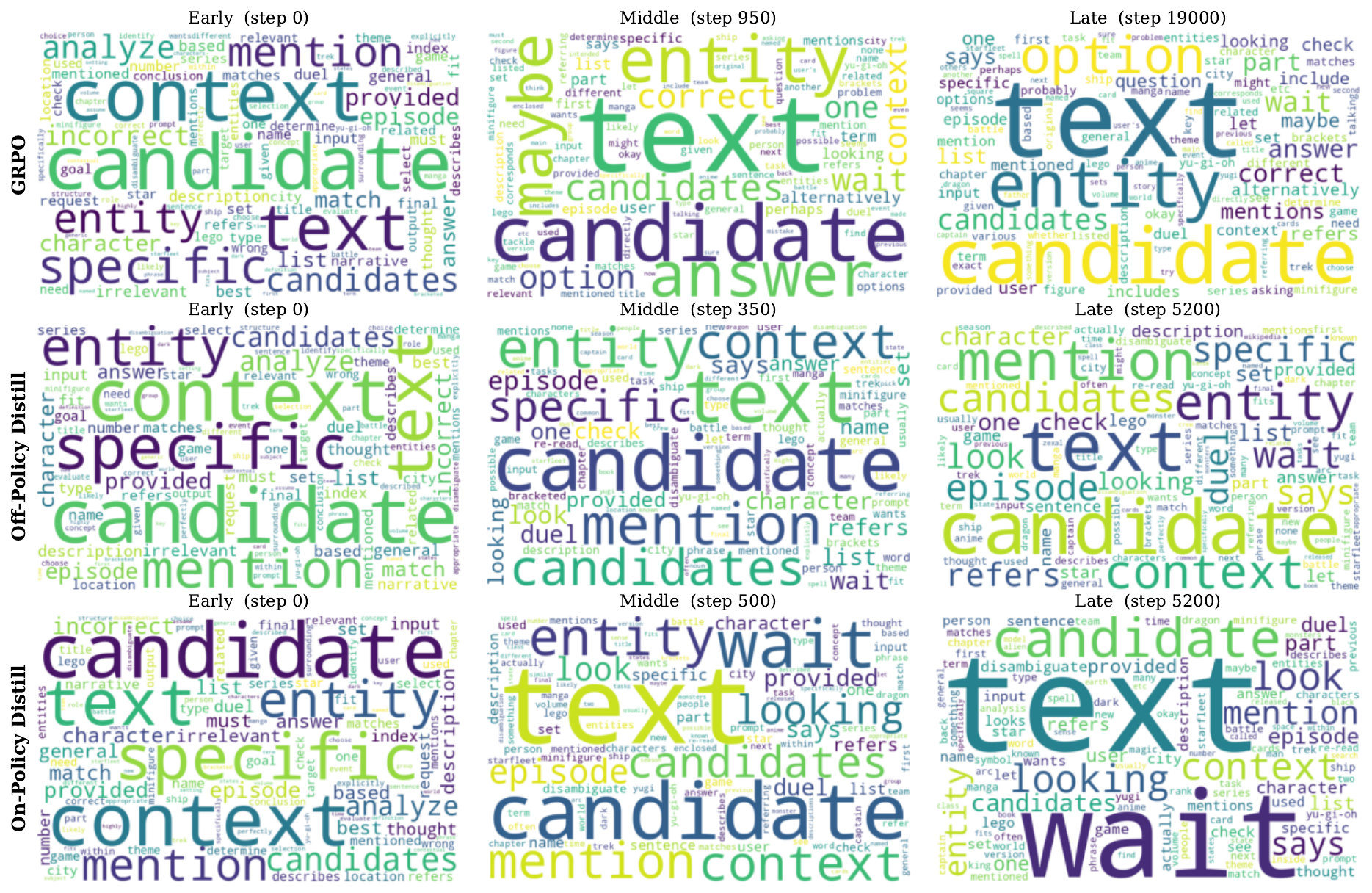}
    \caption{Chain-of-thought word cloud throughout training, using the \texttt{gemma-4-E4B-it} model, pooled across the 250 ZELDA and 4 $\times$ 250 ZESHEL mentions.}
    \label{fig:wordcloud}
    \vspace{-\baselineskip}
\end{figure*}

\section{Training Dynamics}
\label{app:training_dynamics}

\begin{figure*}
  \centering
  \begin{subfigure}{0.49\linewidth}
      \includegraphics[width=\linewidth]{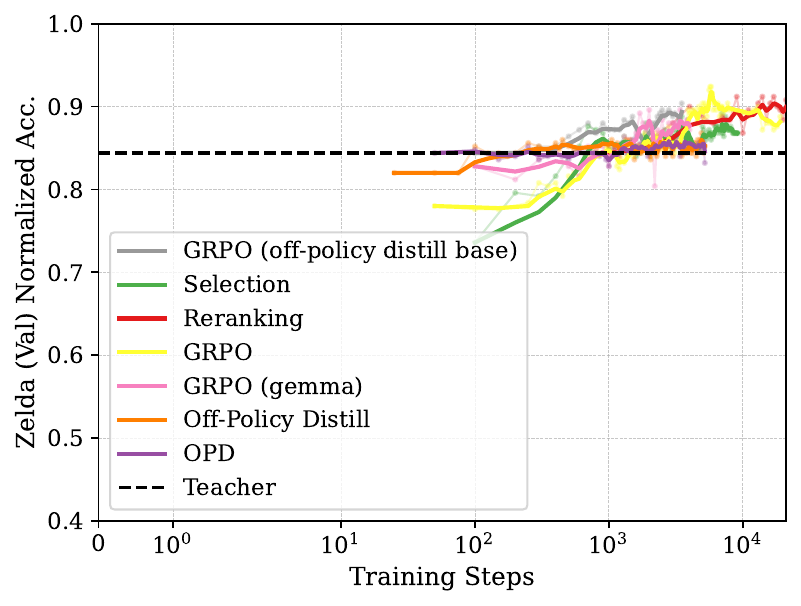}
      \caption{ZELDA, 4B}
      \label{fig:zelda-4b}
  \end{subfigure}
  \hfill
  \begin{subfigure}{0.49\linewidth}
      \includegraphics[width=\linewidth]{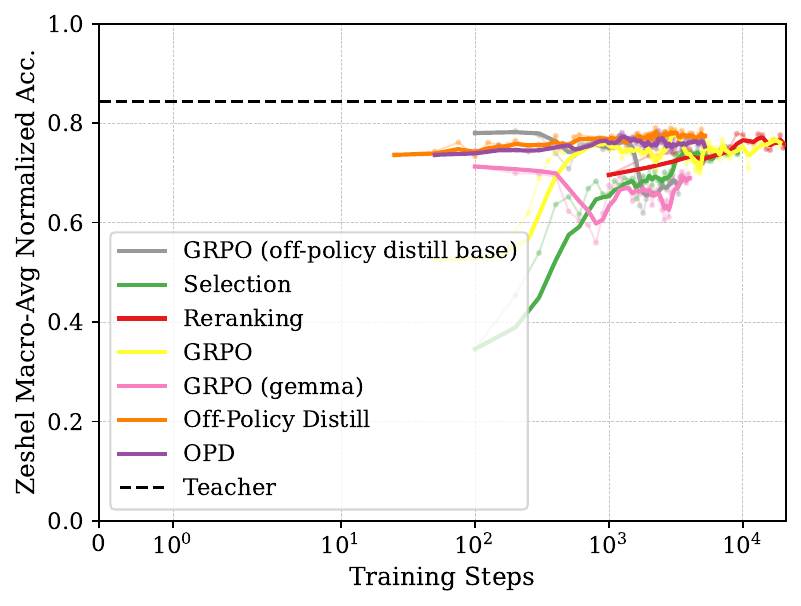}
      \caption{ZESHEL, 4B}
      \label{fig:zeshel-4b}
  \end{subfigure}

  \vspace{0.5em}
  
  \begin{subfigure}{0.49\linewidth}
      \includegraphics[width=\linewidth]{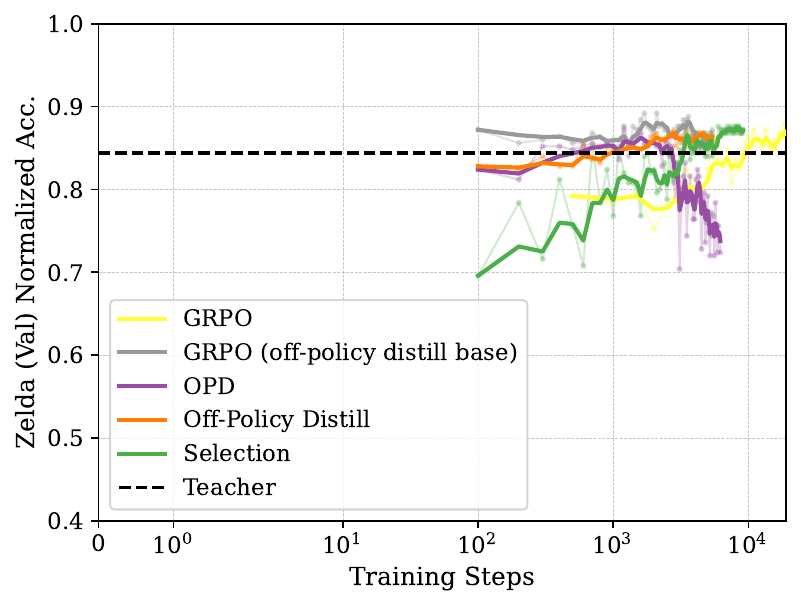}
      \caption{ZELDA, 2B}
      \label{fig:zelda-2b}
  \end{subfigure}
  \hfill
  \begin{subfigure}{0.49\linewidth}
      \includegraphics[width=\linewidth]{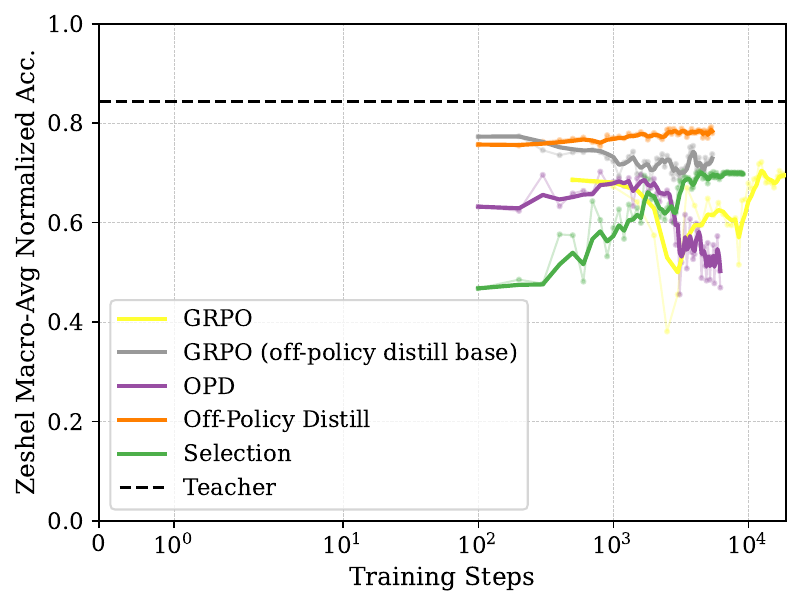}
      \caption{ZESHEL, 2B}
      \label{fig:zeshel-2b}
  \end{subfigure}

  \vspace{0.5em}

  \begin{subfigure}{0.49\linewidth}
      \includegraphics[width=\linewidth]{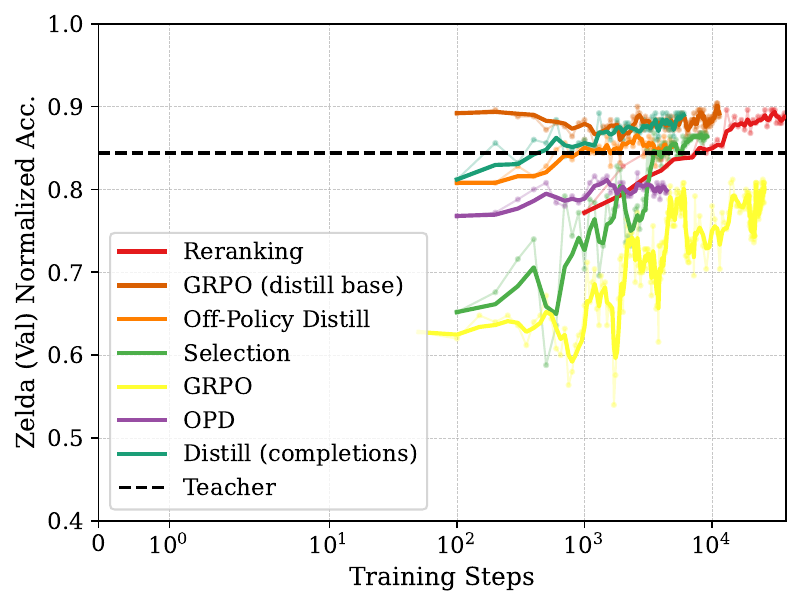}
      \caption{ZELDA, 0.6B}
      \label{fig:zelda-0.6b}
  \end{subfigure}
  \hfill
  \begin{subfigure}{0.49\linewidth}
      \includegraphics[width=\linewidth]{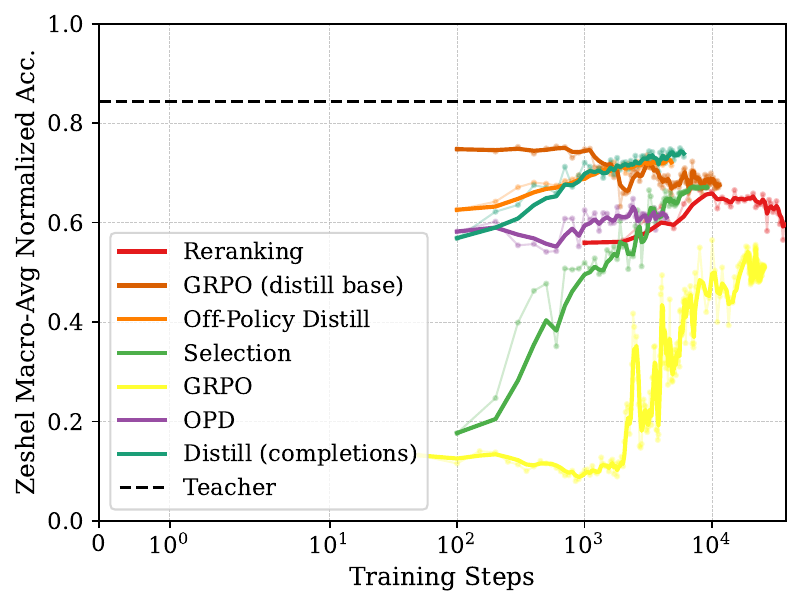}
      \caption{ZESHEL, 0.6B}
      \label{fig:zeshel-0.6b}
  \end{subfigure}
  \caption{Learning curves across paradigms for the different model sizes on ZELDA (val) and ZESHEL (macro-avg).}
  \label{fig:learning-curves}
  \vspace{-\baselineskip}
\end{figure*}

\begin{figure*}
    \centering

    \begin{subfigure}{0.49\linewidth}
        \includegraphics[width=\linewidth]{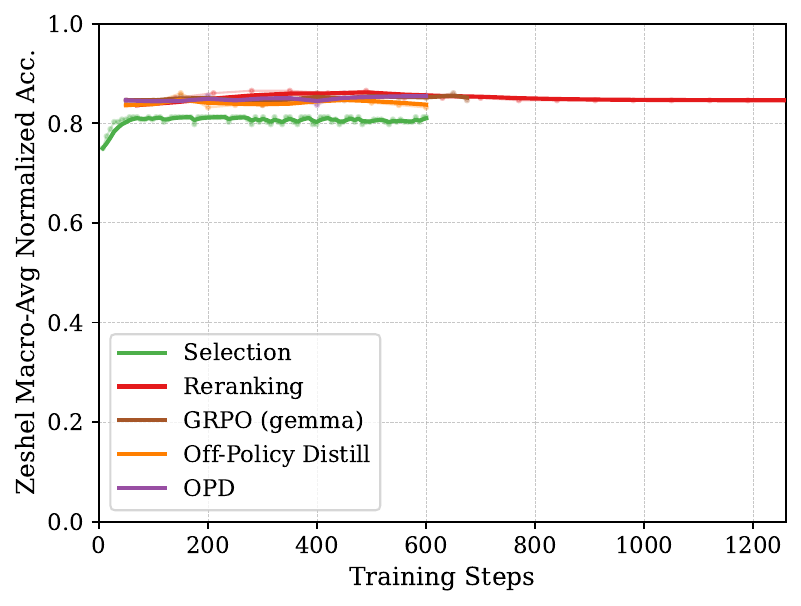}
        \caption{Forgotten Realms}
        \label{fig:fr-4b}
    \end{subfigure}
    \hfill
    \begin{subfigure}{0.49\linewidth}
        \includegraphics[width=\linewidth]{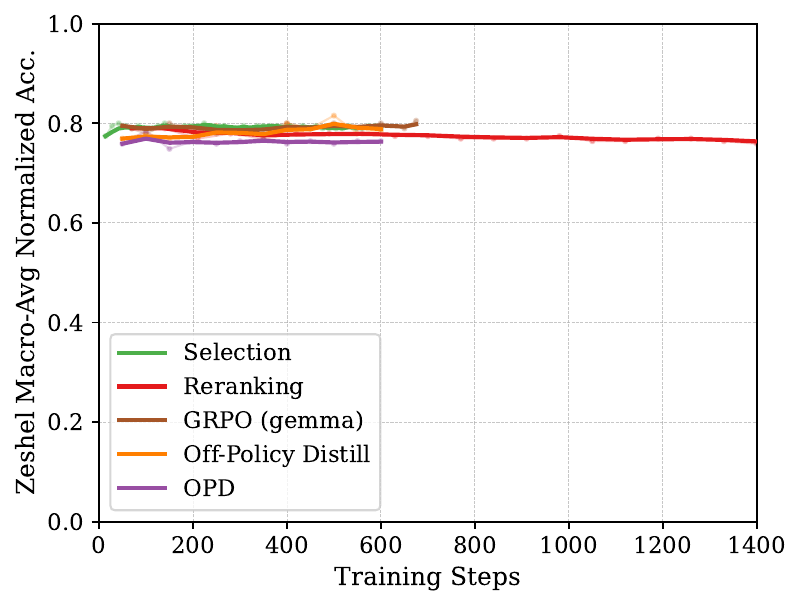}
        \caption{Lego}
        \label{fig:domain2-4b}
    \end{subfigure}

    \vspace{0.5em}

    \begin{subfigure}{0.49\linewidth}
        \includegraphics[width=\linewidth]{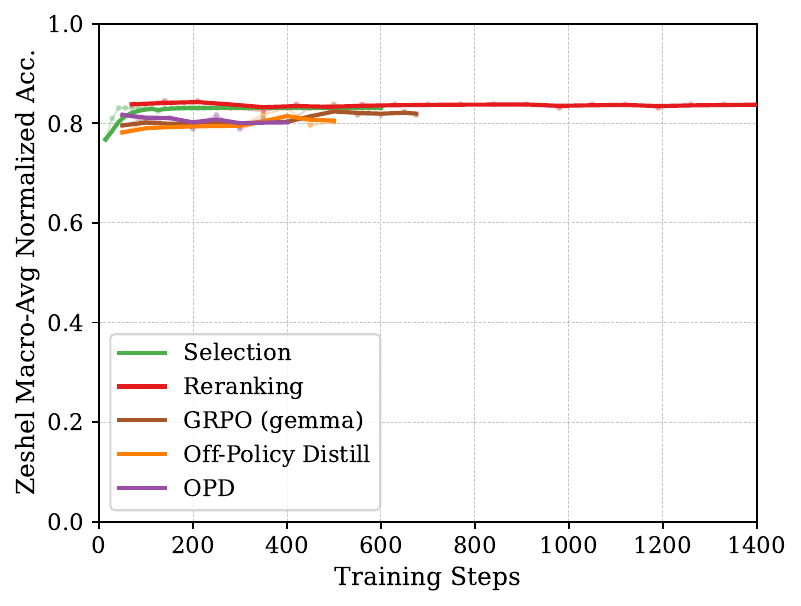}
        \caption{Star Trek}
        \label{fig:domain3-4b}
    \end{subfigure}
    \hfill
    \begin{subfigure}{0.49\linewidth}
        \includegraphics[width=\linewidth]{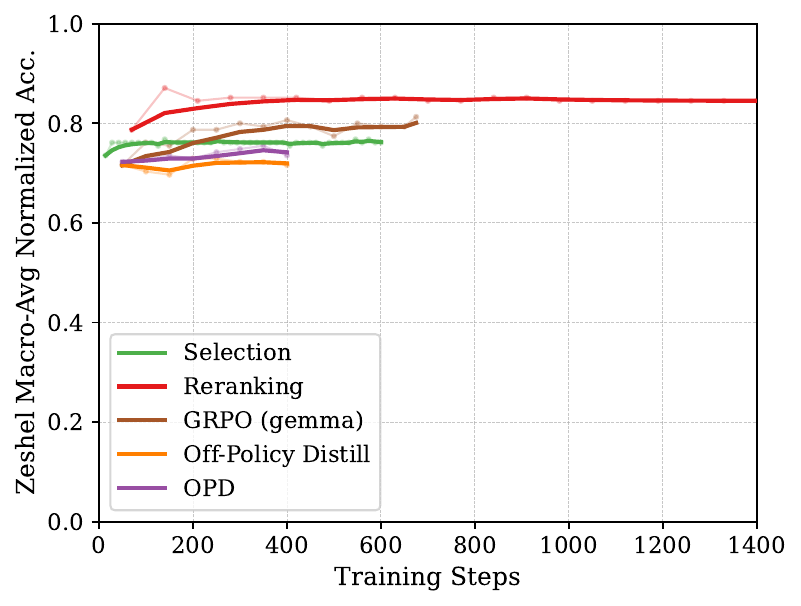}
        \caption{YuGiOh}
        \label{fig:domain4-4b}
    \end{subfigure}

    \caption{Learning curves across paradigms for the 4B model size on four ZESHEL domains.}
    \label{fig:fewshot-4b}
    \vspace{-\baselineskip}
\end{figure*}

\begin{figure*}
  \centering
  \begin{subfigure}{\linewidth}
      \includegraphics[width=\linewidth]{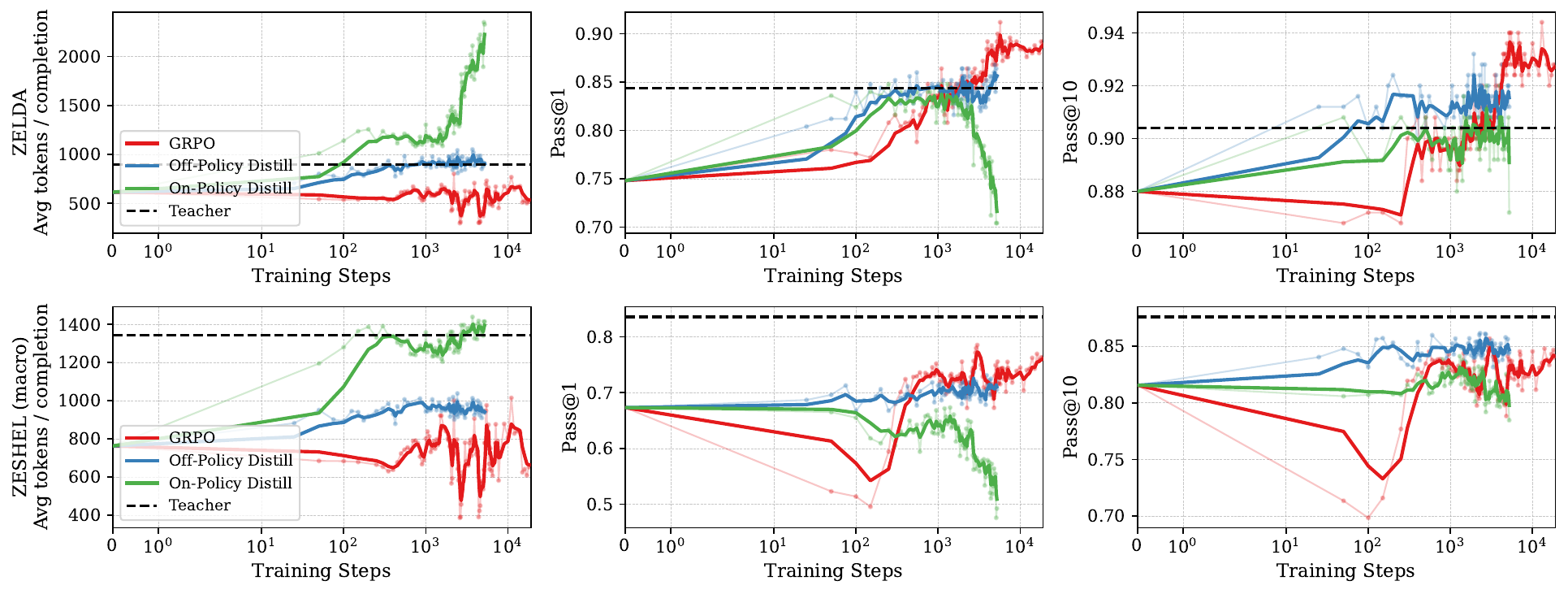}
      \caption{4B}
      \label{fig:sample-stats-4b}
  \end{subfigure}


  \begin{subfigure}{\linewidth}
      \includegraphics[width=\linewidth]{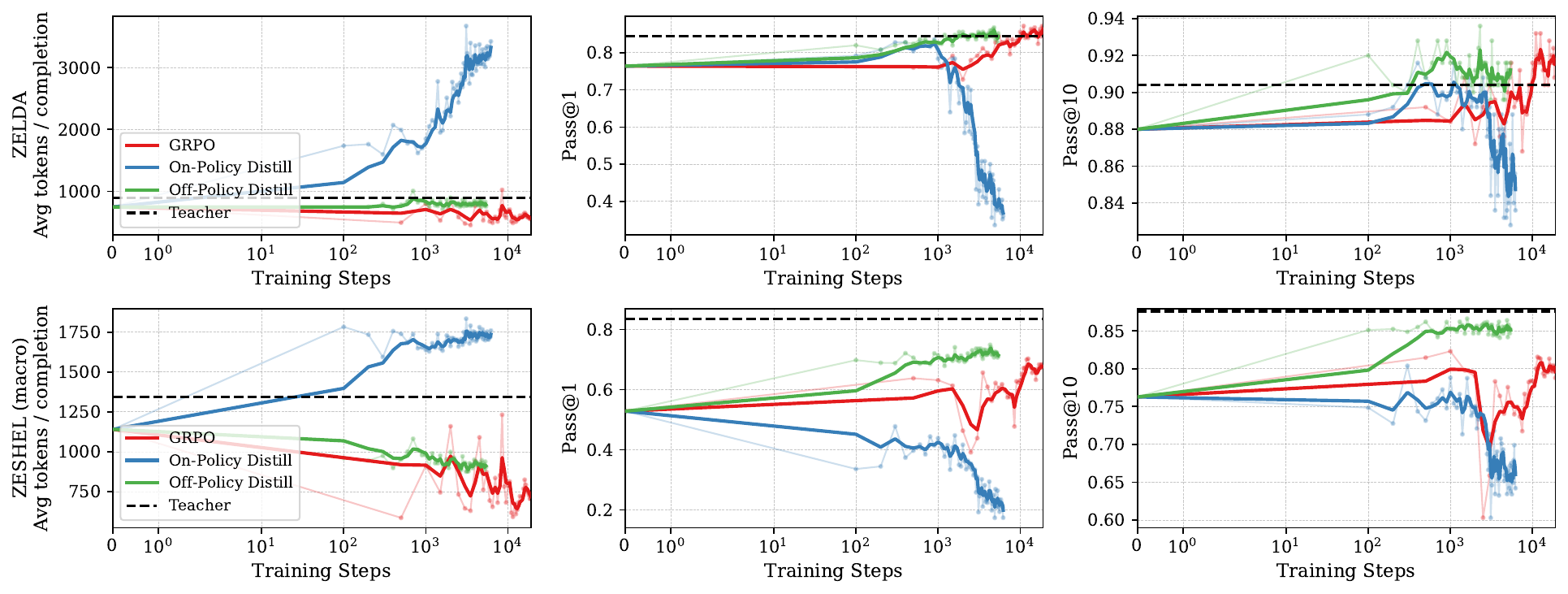}
      \caption{2B}
      \label{fig:sample-stats-2b}
  \end{subfigure}


  \begin{subfigure}{\linewidth}
      \includegraphics[width=\linewidth]{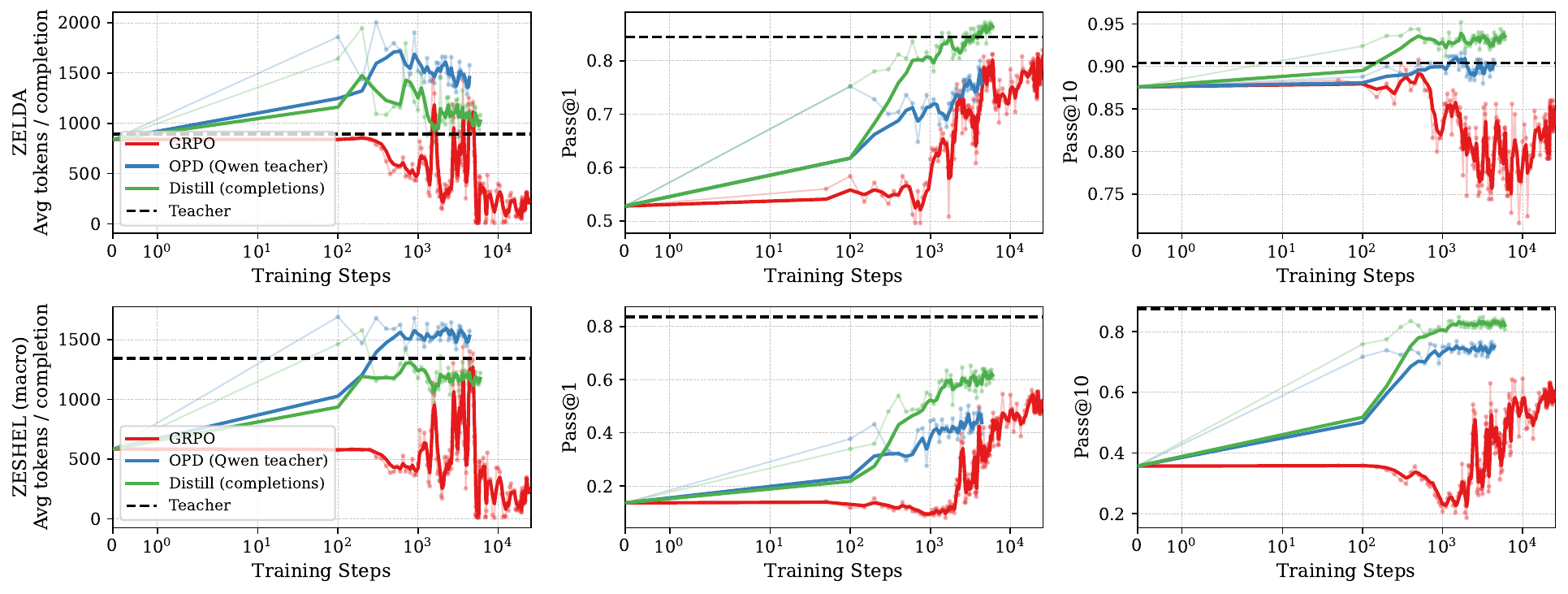}
      \caption{0.6B}
      \label{fig:sample-stats-0.6b}
  \end{subfigure}
  \caption{Evolution of avg.\ output tokens per completion, pass@1, and pass@10 across training steps for GRPO and distillation paradigms. Top row: ZELDA (val). Bottom row: ZESHEL (macro-avg). Dashed line: teacher model reference.}
  \label{fig:sample-stats}
  \vspace{-\baselineskip}
\end{figure*}

\begin{figure*}
  \centering
  \begin{subfigure}{\linewidth}
      \includegraphics[width=\linewidth]{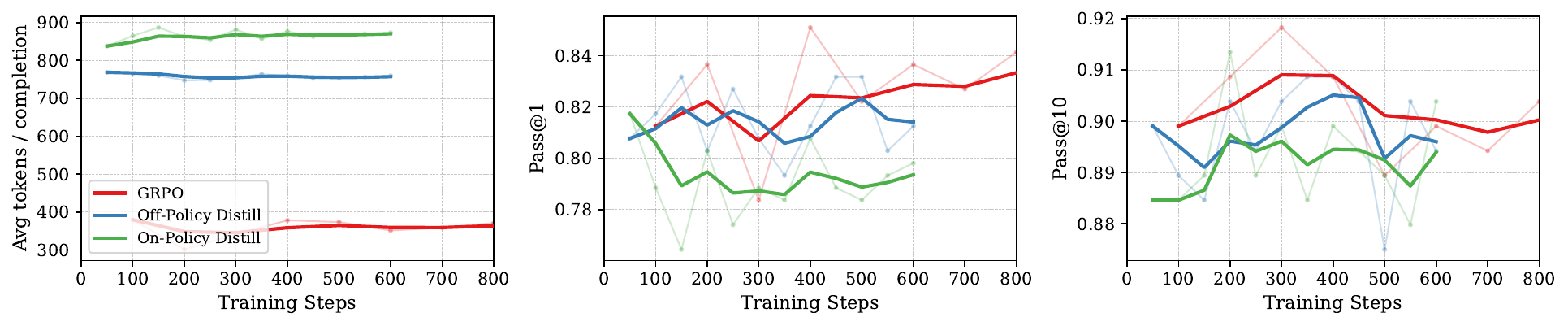}
      \caption{Forgotten Realms}
      \label{fig:sample-stats-4b-fr}
  \end{subfigure}


  \begin{subfigure}{\linewidth}
      \includegraphics[width=\linewidth]{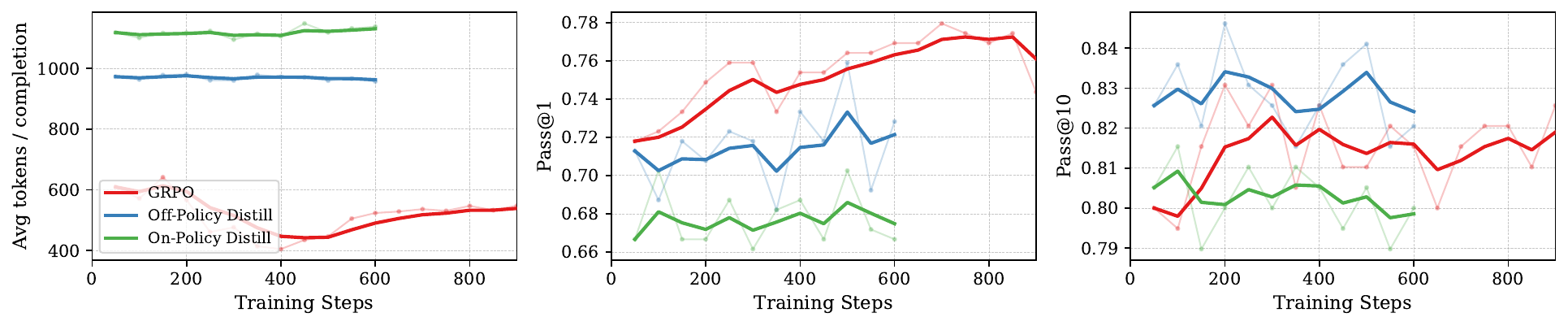}
      \caption{Lego}
      \label{fig:sample-stats-4b-lego}
  \end{subfigure}


  \begin{subfigure}{\linewidth}
      \includegraphics[width=\linewidth]{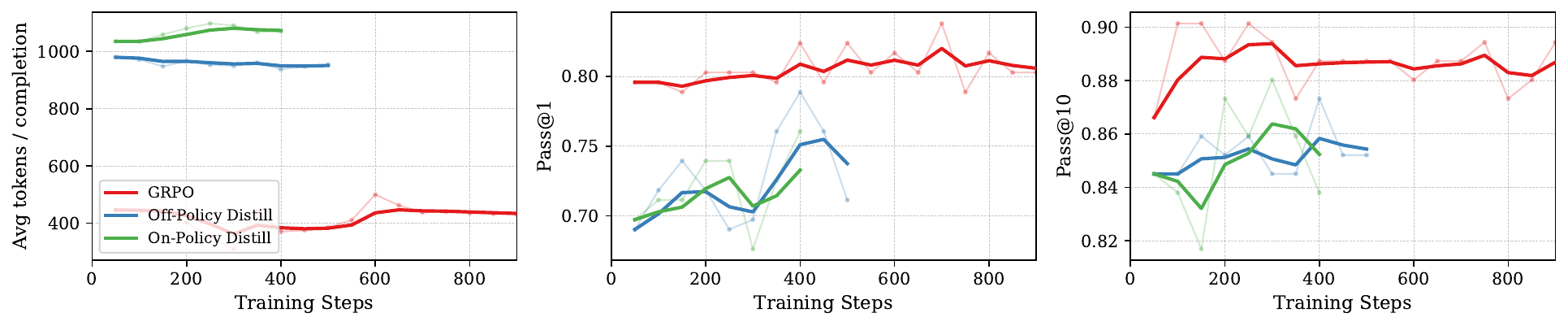}
      \caption{Star Trek}
      \label{fig:sample-stats-4b-star}
  \end{subfigure}
  

  \begin{subfigure}{\linewidth}
      \includegraphics[width=\linewidth]{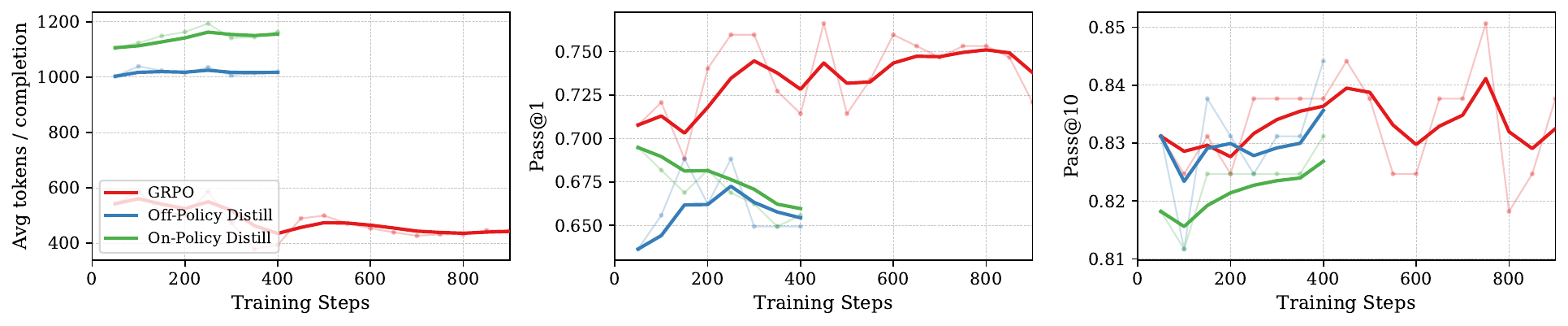}
      \caption{Yugioh}
      \label{fig:sample-stats-4b-yugi}
  \end{subfigure}
  \caption{Evolution of avg.\ output tokens per completion, pass@1, and pass@10 across training steps for GRPO and distillation paradigms for the 4B models, in the few-shot setting on the four ZESHEL test domains.}
  \label{fig:sample-stats-fewshot-4b}
  \vspace{-\baselineskip}
\end{figure*}

We study the training dynamics of the different paradigms under the three ED settings. 
To this end, we sample 250 random mentions from our ZELDA validation split and from each of the four ZESHEL test domains; and evaluate the different model checkpoints on those subsets throughout training.
The learning curves are shown in Figure~\ref{fig:learning-curves} for the classical and true zero-shot settings.
Figure~\ref{fig:fewshot-4b} shows the learning curve for the 4B models on the four domains.
The evolution of Pass@1, Pass@10 and mean output tokens throughout training, in the classical and true zero-shot settings, are shown in Figure~\ref{fig:sample-stats}, and in Figure~\ref{fig:sample-stats-fewshot-4b} in the few-shot domains.

\paragraph{Non-reasoning paradigms.} Reranking converges almost immediately and is flat thereafter: in the few-shot setting it reaches its final accuracy within the first hundred steps and stays there in all four domains (Figure~\ref{fig:fewshot-4b}). Direct selection is the opposite: it starts far below every other paradigm and climbs throughout the run, at every scale (Figure~\ref{fig:learning-curves}), and remains the weakest curve in the few-shot domains.

\paragraph{Reasoning paradigms.} The three training recipes diverge in how
their generations evolve. Under RLVR, trace length falls steadily while
Pass@1 rises towards the base model's Pass@10, consistent with the
sampling-sharpening reading of Appendix~\ref{app:tts}. Both distillation
variants instead lengthen their traces as training proceeds. The on-policy
variant at 2B degenerates, reaching 3{,}000 tokens per completion while
Pass@1 falls\footnote{A preliminary rerun using full-vocabulary distillation showed improved training stability.}.
Figure~\ref{fig:wordcloud} gives the qualitative counterpart. Early in
training all three paradigms produce the same task vocabulary
(\emph{candidate}, \emph{entity}, \emph{context}); by the late checkpoints
the on-policy student is dominated by hedging and backtracking tokens,
\emph{wait} above all, whereas the RLVR student's late traces retain the
vocabulary of the task itself. The on-policy student is scored by a teacher on its own
trajectories, so drifting into ever longer deliberation is never directly
penalised, whereas RLVR's binary reward, though itself indifferent to length, 
shortens traces as the policy concentrates.

\paragraph{GRPO at 0.6B.} At the smallest scale GRPO fails to train stably
(Figure~\ref{fig:learning-curves}e,f): its curves remain erratic and below
direct selection throughout, and the resulting model scores 48.25 on ZESHEL
(Table~\ref{tab:zeshel_results}). This is consistent with RLVR requiring
the base model to reach correct completions often enough for the pass-band
filter of Section~\ref{sec:data} to yield usable signal.

\paragraph{Few-shot adaptation.} Adaptation converges quickly for every paradigm, but not at the same rate: reranking is flat from the first checkpoints, whereas the reasoning recipes take several hundred steps to reach their final accuracy (Figure~\ref{fig:fewshot-4b}).

\section{Forgetting under Domain-Adaptation}
\label{app:continual}

Table~\ref{tab:continual_results} reports accuracy on ZELDA/AIDA after few-shot adaptation to each ZESHEL domain, against the same checkpoint before adaptation. Every paradigm loses some source-domain accuracy, but not to the same degree: RLVR loses between 2.0 and 5.8 points depending on the target domain, whereas both distillation variants stay within 1.4 in every case, and are statistically indistinguishable from the unadapted checkpoint in three out of the four domains. The pointwise reranker and direct selection sit in between.

\begin{table*}
  \centering
  \small
  \begin{tabular}{lccccc}
    \toprule
    \textbf{Method} & $\emptyset$ & \textbf{F. Realms} & \textbf{Lego} & \textbf{Star Trek} & \textbf{YuGiOh} \\
    \midrule
    \multicolumn{6}{c}{\textbf{Pointwise Reranking} $\rightarrow$ \textbf{Pointwise Reranking}} \\
    \midrule
    \texttt{Qwen3-Reranker-4B} & \done{\ci{86.9}{1.0}} & \cellcolor{red!10}\done{\ci{85.6}{1.03}} & \cellcolor{red!10}\done{\ci{85.8}{1.02}} & \cellcolor{red!10}\done{\ci{85.4}{1.03}} & \cellcolor{red!10}\done{\ci{85.5}{1.03}} \\
    \midrule
    \multicolumn{6}{c}{\textbf{Direct Selection} $\rightarrow$ \textbf{Direct Selection}} \\
    \midrule
    \texttt{Qwen3-4B} & \done{\ci{84.7}{1.1}} & \cellcolor{red!15}\done{\ci{82.1}{1.12}} & \cellcolor{red!5}\done{\ci{84.0}{1.07}} & \cellcolor{red!15}\done{\ci{82.6}{1.11}} & \cellcolor{red!15}\done{\ci{82.7}{1.11}} \\
    \midrule
    \multicolumn{6}{c}{\textbf{Selection (Test Time Scaling)}} \\
    \midrule
    \multicolumn{6}{c}{\textbf{Off-Policy Distillation} $\rightarrow$ \textbf{RLVR}} \\
    \texttt{gemma-4-E4B-it} & \done{\ci{87.5}{1.0}} & \cellcolor{red!15}\done{\ci{85.5}{1.03}} & \cellcolor{red!30}\done{\ci{83.8}{1.08}} & \cellcolor{red!25}\done{\ci{84.2}{1.07}} & \cellcolor{red!45}\done{\ci{81.7}{1.13}} \\
    \multicolumn{6}{c}{\textbf{Off-Policy Distillation} $\rightarrow$ \textbf{On-Policy Distillation}} \\
    \texttt{gemma-4-E4B-it} & \done{\ci{87.5}{1.0}} & \cellcolor{red!5}\done{\underline{\ci{87.3}{0.98}}} & \cellcolor{red!10}\done{\ci{86.1}{1.01}} & \cellcolor{red!10}\done{\ci{86.1}{1.01}} & \cellcolor{red!10}\done{\ci{86.2}{1.01}} \\
    \multicolumn{6}{c}{\textbf{Off-Policy Distillation} $\rightarrow$ \textbf{Off-Policy Distillation}} \\
    \texttt{gemma-4-E4B-it} & \done{\ci{87.5}{1.0}} & \cellcolor{red!5}\done{\underline{\ci{87.2}{0.98}}} & \cellcolor{red!10}\done{\ci{86.4}{1.01}} & \done{\underline{\ci{87.5}{0.97}}} & \cellcolor{red!10}\done{\ci{86.2}{1.01}} \\
    \bottomrule
  \end{tabular}
  \caption{Accuracy on ZELDA/AIDA after fine-tuning in the few-shot setting.
  The $\emptyset$ column reports performance before any fine-tuning on ZESHEL.
  Shading gives the relative error increase over that checkpoint,
  $\Delta/(100-\emptyset)$.
  \underline{Underline}: not significantly different from the corresponding
  $\emptyset$ baseline.}
  \label{tab:continual_results}
  \vspace{-\baselineskip}
\end{table*}

\section{Examples}
\label{sec:examples}

\label{app:examples}

Each example below is a uniformly random draw from an
automatically defined subgroup of mentions. All three come from the true
zero-shot setting at 4B, comparing \texttt{Qwen3-Reranker-4B} against
\texttt{gemma-4-E4B-it} trained with off-policy distillation. Recall that the
reasoner selects from the reranker's top-10, so the candidate list it is shown is
the reranker's own ranking; mentions whose gold entity is outside that list are
excluded here. Reasoning traces are excerpted, with elisions marked
\mbox{[\dots]}; the decisive step is set in bold.

\paragraph{Reasoning repairs the reranker.}
Drawn from the $531$ mentions ($7.9\%$) where the reranker's top-1 is wrong, the
gold is its rank-2 candidate, and the reasoner recovers it.
\textsc{lego}, mention \ment{zombies}; gold \texttt{Skeleton Drone}; reranker
\texttt{Mind Control Orb} ($0.0468$) ahead of \texttt{Skeleton Drone} ($0.0464$);
reasoner \texttt{Skeleton Drone} (all $7$ of the $10$ samples that parsed to a
valid candidate).

\begin{quote}\small
[\dots] The game's plot revolves around Ogel making Evil Orbs that turn regular
townspeople into mindless \ment{zombies}, and rescuing captured Alpha Team
members [\dots]
\end{quote}

\begin{quote}\small\itshape
Ogel uses ``Evil Orbs'' to turn ``regular townspeople'' into ``mindless
[zombies]''. [\dots] \dec{Candidate 1 (Mind Control Orb): This is the tool used
to create the zombies. Candidate 2 (Skeleton Drone): This is what the townspeople
become (the minions).} [\dots] Later in the text, it says: ``The zombies were
renamed to skeleton drones'' [\dots] So, ``zombies'' refers to the minions.
Candidate 2 is the most accurate description of these minions.
\end{quote}

\noindent The two candidates are separated by $0.0003$ of reranker score. The
reasoner separates them on a relation the scores cannot express: one entity
causes the state, the other is the state.

\paragraph{Reasoning breaks a correct reranker prediction.}
Drawn from the $106$ mentions ($1.6\%$) where the reranker is right and the
reasoner overturns it with a majority of at least $8/10$.
\textsc{star trek}, mention \ment{performers who were considered}; gold and
reranker top-1 \texttt{Performers considered for Star Trek roles} ($0.961$);
reasoner \texttt{Performers considered for DS9 roles} ($10/10$ samples).

\begin{quote}\small
Performers considered for DS9 roles This is a list of \ment{performers who were
considered} for roles, but ultimately did not appear in the role in the final
episode or film. Performers listed here have been ``verified'' as having been
considered by ``Star Trek'' personnel [\dots]
\end{quote}

\begin{quote}\small\itshape
The context is about ``Performers considered for DS9 roles''. [\dots]
\dec{The text title is ``Performers considered for DS9 roles''.} [\dots]
Candidate 1 is too broad (``Star Trek roles''). \dec{Candidate 2 is specific to
the title and the content provided.}
\end{quote}

\noindent The reasoner treats the title of the document \emph{containing} the
mention as evidence about the mention's referent, and is confidently wrong in all
ten samples.

\paragraph{Both systems fail on the same mention.}
Drawn from the $413$ mentions ($6.2\%$) where both systems are wrong, the
reasoner commits to a majority answer, and the gold is nonetheless inside the
reranker's top-5.
\textsc{yugioh}, mention \ment{the next Rank}; gold
\texttt{Yu - Gi - Oh ! ZEXAL - Rank 041} (reranker rank 4, $0.514$); both systems
answer \texttt{Rank} ($0.973$; reasoner $8$ of $9$ valid samples, the remaining
one answering \texttt{Rank 041}).

\begin{quote}\small
[\dots] Kaito then activates the effect of ``Galaxy Shot'', Tributing ``Full
Armor Photon Dragon''. ``Duel continues in \ment{the next Rank}.'' Featured
cards. The following cards appeared in this chapter. [\dots]
\end{quote}

\begin{quote}\small\itshape
In Yu-Gi-Oh! terminology, ``Rank'' refers to the level/property of Xyz Monsters.
[\dots] \dec{Candidates 3 and 4 are specific chapters, but the text says ``Duel
continues in the next Rank'', not ``Duel continues in Rank 041''.} [\dots] The
entity is a concept within the game mechanics, not a specific character or
chapter title.
\end{quote}

\noindent ``Rank'' is this corpus's word for a manga chapter, so the phrase
refers to the next chapter, \texttt{Rank 041}. The reasoner explicitly considers
the gold and rejects it in favour of the game-mechanics reading; the reranker
makes the same substitution silently. Shared errors are not mentions where the
gold is hidden --- in $43.0\%$ of them it sits at the reranker's rank 2 --- but
mentions where a generic entity is a plausible reading of a specific one. Note
also that one of the ten samples does answer \texttt{Rank 041}: the correct
reading is inside the reasoner's sample distribution and is discarded by the
majority vote, which is consistent with the gap between $\text{pass@}k$ and
majority voting.

\end{document}